\documentclass[lettersize,journal,compsoc]{IEEEtran}

\usepackage{adjustbox}
\usepackage{amsfonts}
\usepackage{amsmath}
\usepackage{booktabs}
\usepackage{colortbl}
\usepackage{forest}
\usepackage[breaklinks=true,colorlinks,citecolor=blue,linkcolor=blue]{hyperref}
\usepackage{gensymb}
\usepackage{graphicx}
\usepackage{fontawesome}
\usepackage{multirow}
\usepackage{tabularx}
\usepackage{tikz} 
\usepackage{pgf, pgfsys, pgffor} 
\usepackage{pgfplotstable} 
\usepackage{pifont}
\usepackage{url}

\pgfplotsset{compat=1.15} 
\usetikzlibrary{trees, positioning, shapes, shadows, arrows.meta} 

\ifCLASSOPTIONcompsoc
  \usepackage[nocompress]{cite}
\else
  \usepackage{cite}
\fi

\newcommand{\eg}{\textit{e.g.}}
\newcommand{\ie}{\textit{i.e.}}

\newcommand{\lnk}[1]{\href{#1}{\faExternalLink}}

\newcommand{\yk}[1]{{\color{black} {#1}}}

\def\acknowledgement{This study is supported by the Ministry of Education, Singapore, under its MOE AcRF Tier 2 (MOE-T2EP20221-0012, MOE-T2EP20223-0002), and under the RIE2020 Industry Alignment Fund – Industry Collaboration Projects (IAF-ICP) Funding Initiative, as well as cash and in-kind contribution from the industry partner(s). (Corresponding author: Ziwei Liu.)
}
\def\affiliations{Y. Cao, F. Hong, Z. Chen, and Z. Liu are with S-Lab, College of Computing and Data Science, Nanyang Technological University, Singapore 639798. E-mail: yukang.cao@ntu.edu.sg, fangzhou.hong@ntu.edu.sg, frozen.burning@gmail.com, ziwei.liu@ntu.edu.sg \\
J. Lu, C. Zhao, Y. Liu are with Intelligent Graphics Lab, The Hong Kong University of Science and Technology. E-mail: lujiahao@mail.ustc.edu.cn, zhaochf.afterjourney@gmail.com, yuanly@ust.hk\\
Z. Shen, Z. Huang, X. Li, and W. Wang are with Texas $A\&M$ University. E-mail: mickshen@tamu.edu, hzs@tamu.edu, xinli@tamu.edu, wenping@tamu.edu\\
}

\begin{document}

\newcolumntype{Y}{>{\centering\arraybackslash}X}
\bstctlcite{IeeeFixDash:BSTcontrol}

\title{Reconstructing 4D Spatial Intelligence: A Survey}

\author{
Yukang Cao, Jiahao Lu, Zhisheng Huang, Zhuowen Shen, Chengfeng Zhao, Fangzhou Hong, Zhaoxi Chen, Xin Li, Wenping Wang, Yuan Liu, Ziwei Liu
\ifCLASSOPTIONcompsoc
  \IEEEcompsocitemizethanks{%
    \IEEEcompsocthanksitem \acknowledgement%
    \IEEEcompsocthanksitem \affiliations%
  }
\else
  \thanks{\acknowledgement}%
  \thanks{\affiliations}%
\fi
}


\IEEEpubid{}
\IEEEtitleabstractindextext{
\begin{abstract}
Reconstructing 4D spatial intelligence from visual observations has long been a central yet challenging task in computer vision, with broad real-world applications. These range from entertainment domains like movies, where the focus is often on reconstructing fundamental visual elements, to embodied AI, which emphasizes interaction modeling and physical realism. Fueled by rapid advances in 3D representations and deep learning architectures, the field has evolved quickly, outpacing the scope of previous surveys. Additionally, existing surveys rarely offer a comprehensive analysis of the hierarchical structure of 4D scene reconstruction. To address this gap, we present a new perspective that organizes existing methods into five progressive levels of 4D spatial intelligence: (1) Level 1 -- reconstruction of low-level 3D attributes (\eg, depth, pose, and point maps); (2) Level 2 -- reconstruction of 3D scene components (e.g., objects, humans, structures); (3) Level 3 -- reconstruction of 4D dynamic scenes; (4) Level 4 -- modeling of interactions among scene components; and (5) Level 5 -- incorporation of physical laws and constraints. We conclude the survey by discussing the key challenges at each level and highlighting promising directions for advancing toward even richer levels of 4D spatial intelligence.
To track ongoing developments, we maintain an up-to-date project page: \url{https://github.com/yukangcao/Awesome-4D-Spatial-Intelligence}.
\end{abstract}

}

\maketitle
\IEEEdisplaynontitleabstractindextext

\ifCLASSOPTIONcompsoc
  {\IEEEraisesectionheading{\section{Introduction}\label{sec:intro}}}
\else
  {\section{Introduction}\label{sec:intro}}
\fi

\IEEEPARstart{T}{he} automatic reconstruction of 4D spatial intelligence using machine learning or deep learning techniques has long been a crucial and challenging problem in computer vision. 
\yk{By capturing both the static configurations and dynamic changes over time, 4D spatial intelligence shall provide a comprehensive representation and understanding of the spatial environments that integrate the three-dimensional geometric structures with their temporal evolution.}
This field has attracted significant attention due to its wide range of applications in video games~\cite{aliev2020neural}, movies~\cite{kim2018deep}, and immersive experiences (AR/VR)~\cite{deng2022fov, li2023instant}, where high-fidelity 4D scenes serve as the foundation for delivering realistic user experiences. 
Beyond these applications that primarily focus on the fundamental components of 4D spatial intelligence -- namely low-level cues such as depth, camera pose, point map, and 3D tracking, as well as scene composing elements and dynamics -- spatial intelligence also plays a pivotal role in advancing embodied AI~\cite{liu2024aligning, gupta2024essential, huang2023embodied} and world models~\cite{zhen20243d}. 
These latter domains place a strong emphasis on the interactions among scene components and the physical plausibility of the reconstructed environments.


In recent years, techniques for reconstructing 4D spatial intelligence have seen rapid advancements. 
Several surveys~\cite{wu2024recent, fei20243d} have provided valuable perspectives from various angles and have highlighted persistent challenges in the field.
For example, \cite{hamid2022stereo, zhou2020review, laga2020survey} reviewed the recent process in deep stereo matching to obtain the low-level scene information;
\cite{gao2022nerf, bao20253d, wang2024nerf} offered a comprehensive overview of advances in 3D scene reconstruction, covering a range of input modalities and diverse 3D representations;
\cite{fei20243d, wu2024recent} classified dynamic 4D scene reconstruction methods into categories based on their core architectural principles.
However, the field has advanced considerably, driven by the emergence of novel 3D representations~\cite{mildenhall2020nerf, shen2021dmtet, kerbl20233d}, high-quality video generation techniques~\cite{ho2022video, ho2022imagen, blattmann2023stable} that provide richer input data, and more efficient models capable of delivering superior reconstruction quality.
Despite these strides, additionally, none of the existing surveys thoroughly examines the different compositional levels of the dynamic 4D scenes, nor do they offer a detailed analysis of their respective developments and open challenges. This would potentially lead to a fragmented understanding that overlooks critical components.
These gaps highlight the need for a comprehensive, up-to-date survey that systematically categorizes 4D spatial intelligence into distinct levels, consolidates recent advancements, and maps the evolving landscape of 4D scene reconstruction.

Driven by this urgent situation, we categorize the existing methods for reconstructing 4D spatial intelligence into five levels and provide a structured overview of their respective advances:

\begin{figure*}[t]
  \centering
  \includegraphics[width=0.92\textwidth]{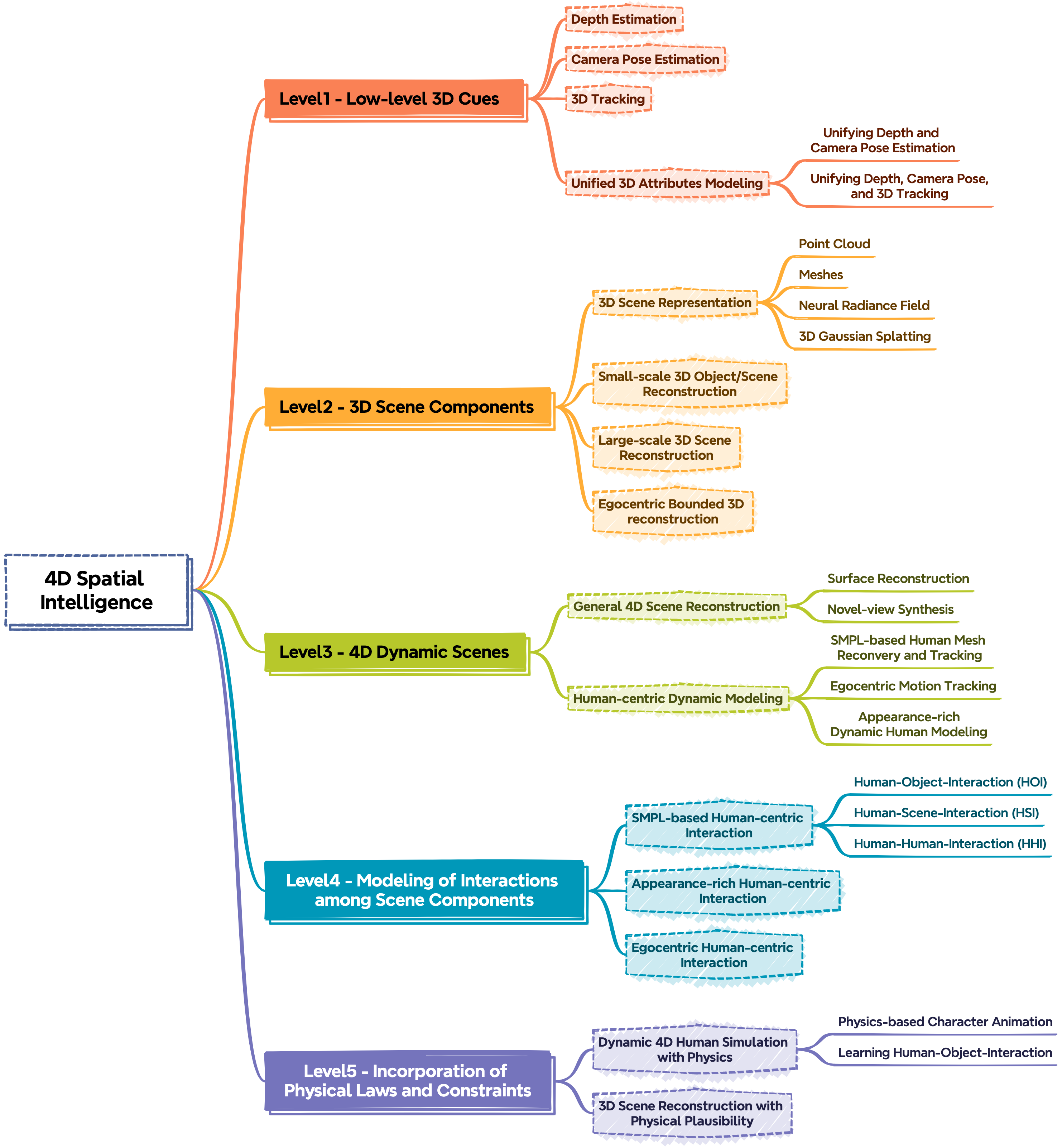}
  \caption{\textbf{Classification of 4D spatial intelligence by level.} Specifically, in this survey, we categorize the methods of reconstructing 3D spatial intelligence from video into five levels: (1) low-level 3D cues, (2) 3D scene components, (3) 4D dynamic scenes, (4) modeling of interactions among scene components, and (5) incorporation of physical laws and constraints.}
  \label{fig:overview_survey}
\end{figure*}                                             
\begin{itemize}
    \item Level 1 -- reconstruction of low-level 3D cues. 
\end{itemize}
At Level 1, the system targets the reconstruction of fundamental 3D cues -- namely, depth, camera pose, point maps, and 3D tracking. These low-level cues define the core structure of a 3D scene. Traditionally, this task has been broken down into separate subfields such as keypoint detection~\cite{ng2003sift,revaud2019r2d2,detone2018superpoint} and matching~\cite{sarlin2020superglue,sun2021loftr,brachmann2019neural,lindenberger2023lightglue}, robust estimation~\cite{barath2023affineglue,brachmann2019neural}, Structure-from-Motion (SfM)~\cite{schonberger2016structure, ozyecsil2017survey,iglhaut2019structure,lindenberger2021pixel}, Bundle Adjustment (BA)~\cite{agarwal2010bundle, engels2006bundle, zach2014robust, triggs1999bundle}, and dense Multi-View Stereo (MVS)~\cite{schoenberger2016mvs,zhang2020visibility,Yang_2020_CVPR,wang2020patchmatchnet,gu2020cascade}. Recent approaches like DUSt3R~\cite{wang2024dust3r} and its follow-ups~\cite{murai2024mast3r, zhang2024monst3r, lu2024align3r, yang2025fast3r} aim to jointly solve these sub-problems, enabling more integrated and collaborative reasoning. Building on transformer-based advances~\cite{han2021transformer, han2022survey, kitaev2020reformer, zhao2021point, parmar2018image}, VGGT~\cite{wang2025vggt} further introduces an end-to-end framework that rapidly estimates these low-level 3D cues within seconds.

\begin{itemize}
    \item Level 2 -- reconstruction of 3D scene components.
\end{itemize}

On top of level 1, level 2 methods move beyond basic 3D cues to reconstruct individual scene elements such as humans, objects, and buildings.
While some methods may involve the composition or spatial arrangement of these elements within a scene, they generally do not model or enforce the correctness of their interactions.
Recent methods for this level leverage the innovations in 3D representations like NeRF~\cite{mildenhall2021nerf}, 3D Gaussians~\cite{kerbl3Dgaussians}, and meshes (\textsc{DMTet}~\cite{shen2021dmtet}, FlexiCube~\cite{shen2023flexicubes}) to improve the reconstructed fine-scale details, rendering efficiency, and global structural coherence, making the results ideal for photorealistic scene reconstruction and immersive virtual experiences.

\begin{itemize}
    \item Level 3 -- reconstruction of 4D dynamic scenes.
\end{itemize}

Level 3 incorporates dynamics into reconstructed 4D scenes, marking a key step toward enabling the “bullet time” experience of 4D spatial intelligence and delivering more immersive visual content. 
Existing approaches can generally be categorized into two main directions. 
The first line of work~\cite{park2021nerfies, tretschk2021non, park2021hypernerf, yu2023dylin, fan2024spectromotion} reconstructs a static canonical radiance field and models temporal changes through learned deformations over time. 
In contrast, the other type of techniques~\cite{li2021neural, gao2021dynamic, you2024decoupling, wang2021neural, lin2023neural, lou2024darenerf, yang2023emernerf} encode time directly as an additional parameter within the 3D representation, allowing for continuous modeling of scene dynamics.

\begin{itemize}
    \item Level 4 -- modeling of interactions among scene components.
\end{itemize}

Advancing beyond the reconstruction of low-level cues, scene elements, and dynamics, Level 4 of spatial intelligence enters a more mature phase -- focusing on modeling interactions between different components within a scene. 
Given that humans are often the central agents of interaction, early works~\cite{dabral2021gravity,xu2021d3d,bhatnagar2022behave,huang2022intercap,jiang2023full} primarily concentrated on capturing the motion of humans and manipulated objects observable in video inputs. 
With recent progress in 3D representations, recent methods~\cite{huo2024stackflow, huo2024monocular, zhao2024imhoi, xie2024sv4d, xie2024intertrack, cao2024avatargo} have achieved more accurate reconstructions of both human and object appearances. 
Furthermore, the study of human-scene interactions~\cite{pavlakos2022one, liu2025josh, zhang2025odhsr, liu2023hosnerf, jiang2022neuman} has gained traction, serving as a foundational step toward constructing comprehensive world models.

\begin{itemize}
    \item Level 5 -- Incorporation of physical laws and constraints.
\end{itemize}

While Level 4 systems are capable of modeling interactions between different scene components, they typically overlook underlying physical principles such as gravity, friction, and pressure. 
As a result, these methods may fall short in applications like embodied AI~\cite{liu2024aligning, gupta2024essential, huang2023embodied}, where the goal is often to enable real-world robots to imitate actions and interactions observed in videos.
Level 5 systems address this limitation by focusing on enforcing physical plausibility within reconstructed 4D scenes. 
Recent approaches~\cite{wang2023physhoi, Luo2023PerpetualHC, luo2024universal}, leveraging platforms such as IsaacGym~\cite{makoviychuk2021isaac} and reinforcement learning techniques~\cite{wiering2012reinforcement, kaelbling1996reinforcement, sutton1999reinforcement}, have demonstrated the ability to learn and replicate human-like skills directly from video data, marking a significant step toward physically grounded spatial intelligence. Beyond human-related applications, the physical modeling of general 3D objects, such as simulating object deformation, collisions, and dynamics, as well as physical scenes, has also become an active area of research~\cite{barhdadi2025physicsnerf, wu2025pbrnerf, yao2025cast}, expanding the scope and applicability of Level 5 reconstruction systems.



\noindent \textbf{Scope.} 
This survey primarily focuses on approaches for reconstructing 4D scenes from video inputs. 
Specifically, we examine key developments and representative works across our defined Levels 1 through 5 of the 4D sptial intelligence.
The papers reviewed are predominantly drawn from leading conferences and journals in computer vision and computer graphics, along with select preprints released on arXiv in 2025. Our selection criteria emphasize relevance to the scope of this survey, with the goal of providing a comprehensive overview of recent rapid progress in the field.
We do not include the 3D generation methods~\cite{voleti2025sv3d, chen2024v3d, cao2024dreamavatar} and 4D generation approaches~\cite{bahmani20244d, bahmani2024tc4d, zhang20244diffusion, liang2024diffusion4d, cao2024crowdmogen, cao2023guide3d} based on generative video diffusion models~\cite{ho2022video, ho2022imagen, blattmann2023stable}, as these methods typically yield a single type of input and have limited direct relevance to 4D reconstruction techniques. Additionally, this survey does not delve into a detailed analysis of various 3D representations. Readers interested in these complementary areas are encouraged to read existing surveys on 4D generation~\cite{miao2025advances, li2023generative, liu2024comprehensive, li2024advances} and the evolution of 3D representation methods~\cite{bao20253d, fei20243d, kato2020differentiable, deng2022survey}.

\noindent \textbf{Organization.} 
An overview of the different levels of 4D spatial intelligence is illustrated in Fig.~\ref{fig:overview_survey}. 
In the following sections, we introduce a taxonomy that organizes recent research efforts according to the evolving process of reconstructing five key levels from video inputs: low-level 3D cues (Sec.~\ref{sec:level1}), basic 3D scene components (Sec.~\ref{sec:level2}), dynamic 4D scenes (Sec.~\ref{sec:level3}), interaction between scene components (Sec.~\ref{sec:level4}), and physics modeling (Sec.~\ref{sec:level5}). 
The overall structure of the survey is summarized in Fig.~\ref{fig:overview_survey}. 
Finally, in Sec.~\ref{sec:challenges}, we critically reflect on current methodologies, identify open challenges at each level of spatial intelligence, and discuss future directions for advancing 4D spatial intelligence beyond these five defined levels.
\section{Level 1 -- Low-level 3D cues}
\label{sec:level1}

\begin{figure}[t]
  \centering
  \includegraphics[width=\linewidth]{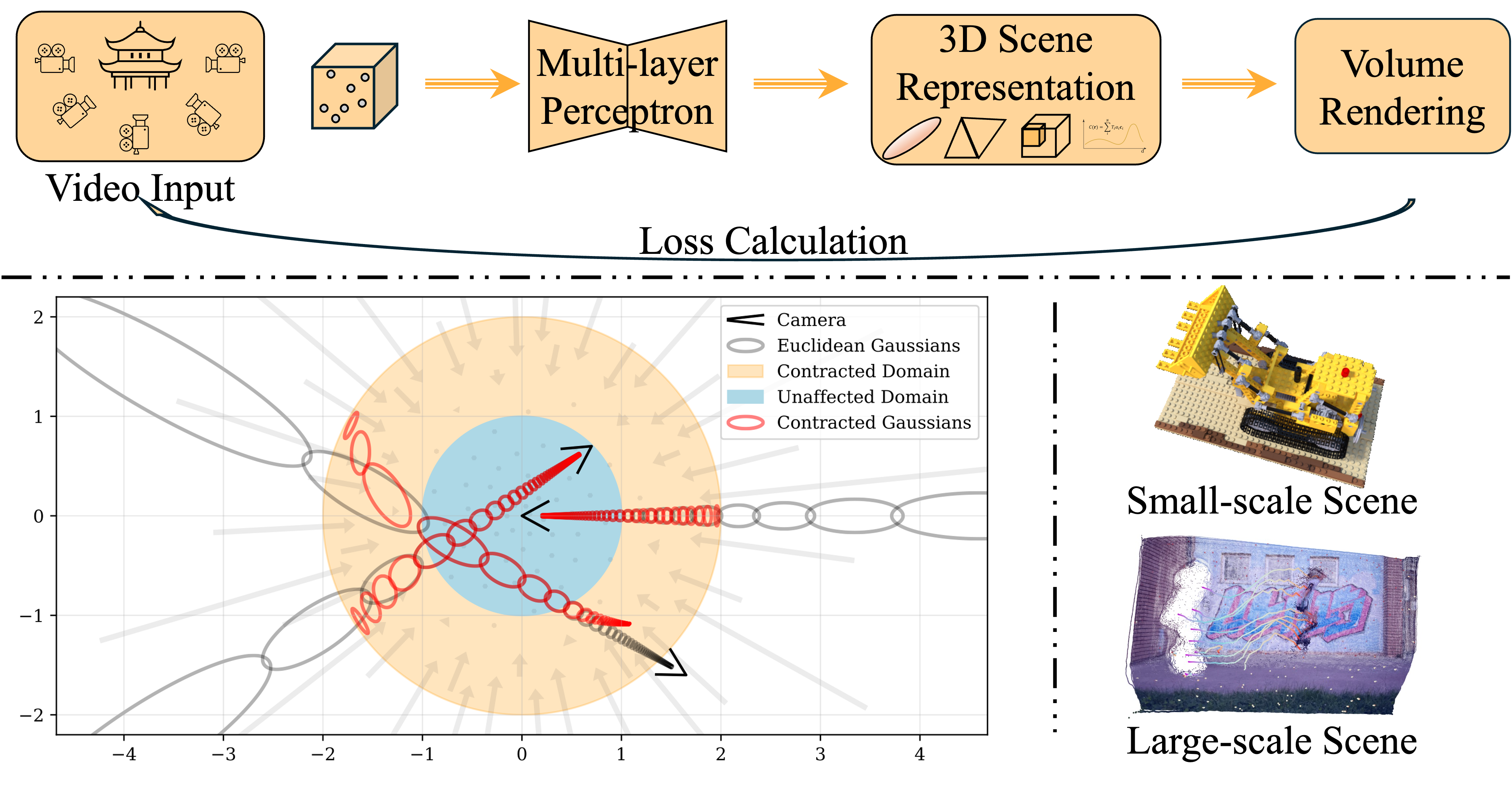}
  \caption{\textbf{The paradigms of methods for reconstructing 3D scene components from video input.} 3D reconstruction methods for small-scale and large-scale scenes often share similar architectures, differing primarily in the spatial extent they handle. As shown in the left panel (Image source: MipNeRF360~\cite{mipnerf3602022}), small-scale scenes correspond to the unaffected domain. large-scale scenes additionally incorporate a contracted domain. Examples illustrating both scene types are provided in the right panel.}
  \label{fig:level1}
\end{figure}

Depth, camera pose, and 3D tracking are commonly regarded as low-level cues in 3D scene modeling. 
These parameters capture the fundamental geometric and positional structure of the environment, forming the basis for higher-level tasks such as object reconstruction, scene composition, and physical interaction modeling. 
In this sense, they function similarly to pixels and edges in 2D vision. 
As such, we define the reconstruction of these elements as level 1 of 4D spatial intelligence.
The paradigms of the methods for obtaining these low-level cues from videos are illustrated in Fig.~\ref{fig:level1}. They can be further categorized according to their respective objectives and the type of input videos.

\subsection{Depth estimation}

Video-based depth estimation aims to generate accurate and temporally consistent depth maps from RGB video sequences. Early approaches typically relied on inference-time optimization to align depth across frames~\cite{pacheco2020inference}, or employed self-supervised warping using estimated ego-motion and optical flow~\cite{yin2018geonet, gordon2019depth, bian2019unsupervised}, often further enhanced by test-time refinement~\cite{luo2020consistent, chen2019self}. While effective, these methods are computationally expensive and heavily dependent on the accuracy of pose and flow estimations.
To address these challenges, feed-forward architectures have been introduced. Cost-volume–based methods construct 3D matching volumes to enforce temporal coherence~\cite{long2021multi, guizilini2022multi, watson2021temporal, sayed2022simplerecon}, while flow-guided approaches integrate optical flow cues directly~\cite{xie2020video, eom2019temporally}. Recurrent models leverage temporal recurrence to iteratively refine predictions across frames~\cite{patil2020don, zhang2019exploiting}, and attention-based mechanisms dynamically reweight temporal features~\cite{wang2022less, yasarla2023mamo, zhao2022monovit}. Other notable feed-forward systems include~\cite{li2023temporally, Teed2020DeepV2D, NVDS, NVDSPLUS, yasarla2023mamo}.
More recently, large-scale pretraining and diffusion-based frameworks have pushed the frontier further. DepthCrafter~\cite{hu2024depthcrafter}, ChronoDepth~\cite{shao2024chronodepth}, and DepthAnyVideo~\cite{yang2024depthanyvideo} leverage video diffusion models to generate depth sequences directly, while \cite{chen2025video} extends the ViT-based Depth Anything V2~\cite{yang2024depthanythingv2} for video depth estimation. These models exhibit strong temporal consistency and robust generalization across diverse scenes.
Overall, the field has progressed from optimization-heavy, pose-dependent pipelines to efficient feed-forward networks, and most recently, to pretrained, diffusion-driven models that achieve both high accuracy and temporal coherence.

\begin{figure}[t]
  \centering
  \includegraphics[width=\linewidth]{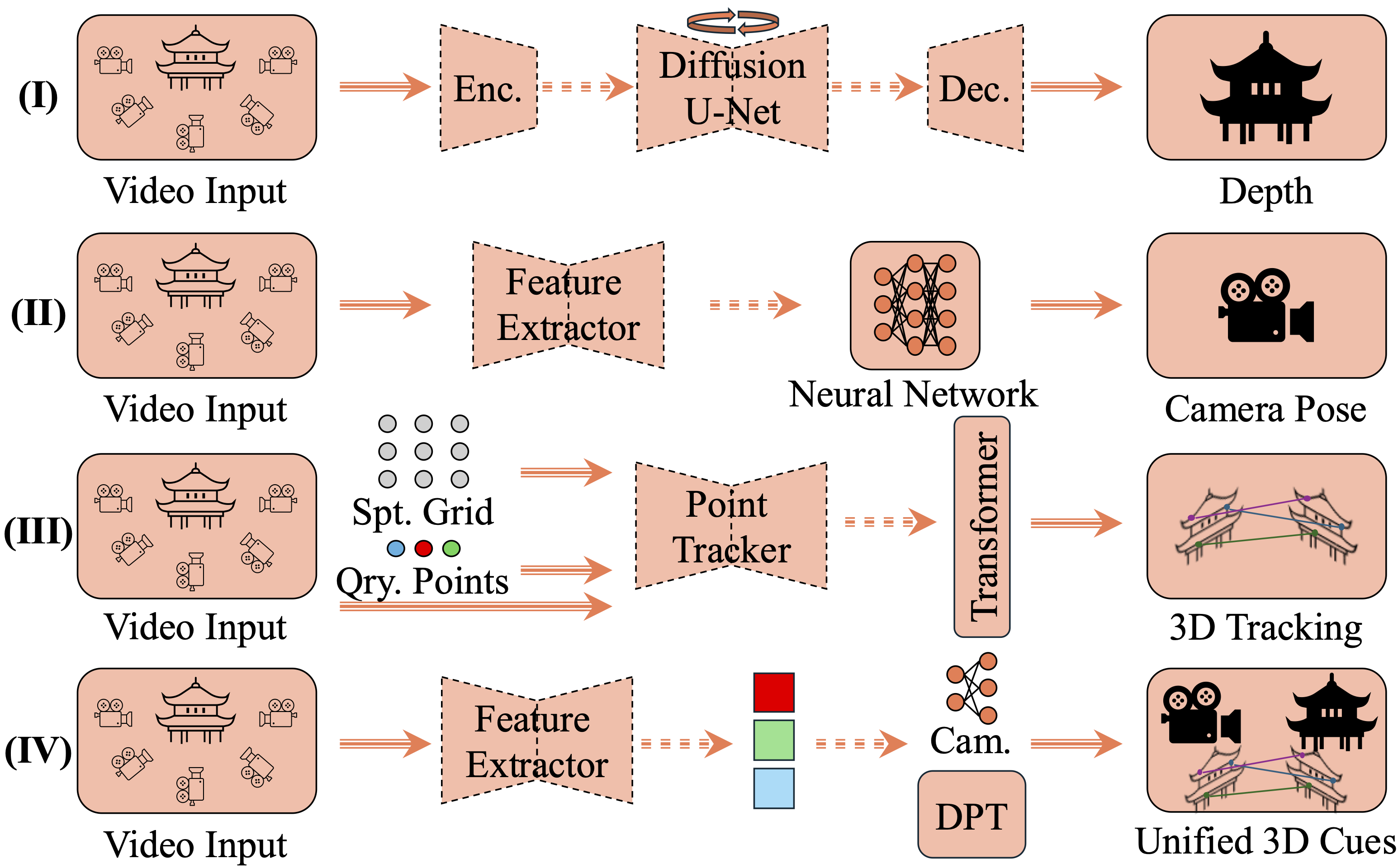}
  \caption{\textbf{The paradigms of methods for reconstructing low-level cues from video input.} (I) Video-based depth reconstruction methods recently leverage the diffusion model to obtain the depth maps; (II) Methods for reconstructing camera pose from video input typically employ the neural network to infer the camera pose based on the encoded image features; (III) 3D tracking methods uses point tracker and transformers to achieve 3D tracking from video input; (IV) Recent methods, such as VGGT, apply DINO to extract the features and then train transformer-based DPT heads to infer the unified 3D attributes. ``Enc.'', ``Dec.'', ``Spt. Grid'', ``Qry. Points'', and ``Cam.'' denote ``Encoder'', ``Decoder'', ``Supporting Grid'', ``Query Points'', and ``Camera Head'' correspondingly.}
  \label{fig:level2}
\end{figure}
\subsection{Camera pose estimation}
Camera pose estimation from RGB videos can be generally solved by Visual Odometry (VO) algorithms, which are widely applied in robotics applications.
Classical geometry-based VO methods are typically categorized into two groups: feature-based and direct approaches.
Feature-based VO~\cite{mur2017orb, wang2021line, kannapiran2023stereo, shu2023structure, jiang2024ul} estimates camera motion by detecting and tracking visual features across frames, while direct VO~\cite{engel2014lsd, engel2017direct, zhou2022edplvo, yang2020d3vo} infers motion by minimizing photometric error or applying feature warping~\cite{pelecanos2001feature}.
With the advent of deep learning, learning-based VO methods~\cite{wang2021tartanvo, wang2017deepvo, shen2023dytanvo} have gained prominence, often outperforming traditional approaches in controlled settings but facing challenges in generalizing to unseen environments.
To overcome these limitations, hybrid methods~\cite{zhao2021deep, teed2024deep, xu2025airslam, lipson2024deep, lai2023xvo, wimbauer2025anycam, rockwell2025dynamic} have been proposed to combine learning-based techniques with geometric insights, leveraging the strengths of both paradigms.
More recently, to reduce reliance on manual hyperparameter tuning required by these hybrid methods, further studies~\cite{zhang2024efficient, messikommer2024reinforcement} have explored reinforcement learning for adaptive decision-making in VO systems.
It is also worth noting that VO is closely related to Visual Simultaneous Localization and Mapping (VSLAM), which extends VO by concurrently constructing a map of the environment.
Methods that jointly estimate camera pose and dense depth for mapping purposes will be discussed in a later section on unified camera pose and depth estimation from video.

\subsection{3D tracking}
3D tracking estimation aims to recover the motion of scene elements in dynamic videos, providing temporally coherent correspondences in 3D space.
A notable approach in this area is OmniMotion~\cite{wang2023tracking}, which represents an input video using a quasi-3D canonical volume and performs dense, pixel-wise tracking by establishing bijective mappings between the local input space and the canonical space.
Through per-video optimization, it jointly estimates the motion trajectories across the entire sequence, enabling consistent tracking over time.
Building upon this framework, OmniTrackFast~\cite{song2024track} enhances both computational efficiency and robustness by factorizing the underlying function representation into a local spatiotemporal feature grid, and further improves the model’s expressiveness by introducing non-linear functions into the coupling blocks.
In contrast to these optimization-heavy methods, SpatialTracker~\cite{xiao2024spatialtracker} proposes a feed-forward architecture that supports long-range 3D tracking across videos without the need for test-time optimization, offering a more scalable and efficient alternative. 
SceneTracker~\cite{wang2024scenetracker} employs an iterative strategy to approximate the optimal 3D trajectory, dynamically indexing and constructing both appearance correlation and depth residual features in parallel.
DELTA~\cite{ngo2024delta} introduces a coarse-to-fine trajectory estimation strategy, allowing for efficient dense tracking across the entire frame rather than being limited to a sparse set of locations. 
Seurat~\cite{cho2025seurat} derives depth directly from 2D tracking inputs to recover 3D trajectories.
TAPIP3D~\cite{zhang2025tapip3d} constructs spatio-temporal feature clouds from videos by utilizing depth and camera motion information to project 2D video features into a 3D world space, where the effects of camera movement are effectively neutralized.
Recent methods, such as EgoPoints~\cite{darkhalil2025egopoints}, introduces a new benchmark and new metrics for point tracking from egocentric videos. It opens the door for future works.
Together, these methods illustrate the evolving landscape of 3D tracking, spanning from optimization-based pipelines to fully end-to-end learning systems.

\subsection{Unified 3D cues modeling}

Accurate 4D scene reconstruction requires not only high-quality geometry but also coherent spatial-temporal understanding across frames. To achieve this, recent research has moved toward unified frameworks that jointly estimate depth, camera pose, and even 3D tracking from video. These joint approaches reduce the inconsistencies and ambiguities that often arise when these components are estimated independently, leading to more robust and temporally consistent reconstructions. The field has seen a wide range of solutions—from optimization-based pipelines that refine predictions per video, to end-to-end feed-forward architectures designed for efficiency and scalability. In this section, we review representative methods that integrate one or more of these key components, highlighting their methodological designs, trade-offs, and contributions to the broader goal of learning-based 4D spatial intelligence.

\noindent \textbf{Unifying depth and camera pose estimation}
Jointly predicting depth and camera pose from monocular video is a foundational step toward full 4D scene reconstruction. 
Many recent methods tackle this challenge by leveraging monocular depth priors and applying per-video optimization to enforce temporal and geometric consistency.
For example, Robust-CVD~\cite{kopf2021robust} applies flexible deformation splines for large-scale geometric alignment and introduces geometry-aware filtering to refine high-frequency depth details;
CasualSAM~\cite{zhang2022structure} fine-tunes a monocular depth model on individual video sequences to jointly optimize both depth and camera pose. 
MegaSaM~\cite{li2024megasam} adapts traditional visual SLAM paradigms to handle dynamic scenes, maintaining dense map construction and pose estimation. 
More recently, DUSt3R~\cite{wang2024dust3r} proposes a unified framework that simultaneously predicts depth, camera pose, and a dense point map, allowing mutual refinement across these outputs.
Building on this idea, MonST3R~\cite{zhang2024monst3r} and Align3R~\cite{lu2024align3r} extend the approach to dynamic scenes by fine-tuning on motion-rich datasets and producing temporally consistent point trajectories.
Align3R further integrates monocular depth priors to enhance reconstruction quality, although both methods still rely on global alignment during post-processing.
Easi3R~\cite{chen2025easi3r} is a simple yet efficient training-free method for 4D reconstruction. It adapts attention during inference, eliminating the need for pre-training from scratch or network fine-tuning. GeometryCrafter~\cite{xu2025geometrycrafter} introduces a pointmap Variational Autoencoder (VAE) that learns a latent space independent of video-specific distributions, enabling effective pointmap encoding and decoding for accurate 3D/4D reconstruction and camera parameter estimation.
In contrast to optimization-heavy pipelines, Spann3R~\cite{wang20243d} adopts a feed-forward approach that enables continuous 4D reconstruction via a spatial memory mechanism. CUT3R~\cite{wang2025continuous}, on the other hand, leverages a compressed state representation that not only encodes observed information but also supports the inference of unobserved structures.
Point3R\cite{point3r} and StreamVGGT\cite{streamVGGT} further enhance the capabilities of streaming 3D/4D reconstruction.
Pi3~\cite{wang2025pi3} introduces an innovative feed-forward neural network that fundamentally changes visual geometry reconstruction by removing the dependency on a fixed reference view.
Several diffusion-based methods~\cite{team2025aether, jiang2025geo4d, sun2025unigeo} have also demonstrated strong performance on this task. Aether\cite{team2025aether}, Geo4D\cite{jiang2025geo4d}, and UniGeo~\cite{sun2025unigeo} can simultaneously predict high-quality depth and accurate camera poses, benefiting from their denoising-based designs.

\noindent \textbf{Unifying depth, camera pose, and 3D tracking}
Recent methods have made significant progress toward jointly estimating video depth, camera pose, and 3D tracking. 
Uni4D~\cite{yao2025uni4d} adopts a multi-stage optimization framework that integrates multiple pretrained models to handle both static and dynamic 3D reconstruction. 
BA-Track~\cite{chen2025back} disentangles camera-induced motion from object motion using a 3D point tracker, enabling robust bundle adjustment across the entire scene, and further enforces temporal depth consistency through scale-map-based post-processing.
Built upon DUSt3R~\cite{wang2024dust3r}, several recent works extend its capability toward dynamic 3D reconstruction and tracking.
Stereo4D~\cite{jin2024stereo4d} leverages large-scale training on temporally consistent 3D point clouds to recover long-term pseudo-motion trajectories along with depth and camera pose.
DPM~\cite{sucar2025dynamic} introduces time into the representation, leading to multiple possible spatial-temporal reference frames for defining point maps. The authors identify a minimal and sufficient subset of these reference combinations that can be regressed by a network to address the aforementioned sub-tasks.
St4RTrack~\cite{feng2025st4rtrack} jointly predicts point maps at a single timestamp within a unified world coordinate system and chains these predictions across time to reconstruct long-range trajectories, effectively integrating reconstruction and tracking.
POMATO~\cite{zhang2025pomato} combines pointmap matching with temporal motion modeling, establishing cross-view pixel-to-pointmap correspondences and introducing a temporal module to enforce scale consistency and improve 3D point tracking.
D2USt3R~\cite{han2025d} directly regresses 4D pointmaps that represent both static and dynamic scene geometry, explicitly modeling spatial and temporal aspects to provide dense spatio-temporal correspondences beneficial for downstream tasks.
Zero-MSF~\cite{liang2025zero} presents a joint geometry-and-motion estimation framework supported by a large-scale data generation pipeline, which produces 1M annotated samples from diverse synthetic scenes. It also identifies and adopts an effective parameterization for scene flow through systematic evaluation.
In contrast to the optimization-heavy methods above, TracksTo4D~\cite{kastenfast} operates in an unsupervised and feed-forward manner. It takes 2D point tracks as input and predicts full 4D structures from casually captured videos without requiring ground truth or supervision. Leveraging recent advancements in transformers~\cite{khan2022transformers}, VGGT~\cite{wang2025vggt} introduces an end-to-end architecture capable of efficiently predicting low-level 3D cues within seconds. SpatialTrackerV2~\cite{xiao2025spatialtrackerv2} proposes a unified, feedforward 3D point tracker that integrates point tracking, monocular depth, and camera pose estimation. It decomposes world-space motion into scene geometry, camera ego-motion, and pixel-wise object motion, with a fully differentiable, end-to-end architecture that scales across synthetic, RGB-D, and in-the-wild data.

On the other hand, jointly estimating depth and camera pose from dynamic RGB video remains an open challenge due to occlusions, object motion, and scene complexity.
Several methods address this by enforcing temporal and geometric consistency within short frame windows~\cite{li2018undeepvo, teed2018deepv2d, li2021generalizing, zhao2021deep, sun2022improving, zhao2020towards}, yielding locally consistent results.
However, these methods often suffer from accumulated errors over longer sequences. 
In response, dense visual SLAM methods~\cite{lu2023deep, teed2021droid, zhu2024nicer, li2023dense} extend traditional SLAM pipelines to produce globally consistent and dense depth maps instead of sparse point clouds. 
Notably, recent SLAM systems~\cite{zhang2023go, zhang2024glorie, sandstrom2024splat, matsuki2024gaussian} adopt 3D Gaussian Splatting~\cite{kerbl20233d} as a scene representation, benefiting from its real-time rendering and high-fidelity geometry. 
Please refer to the survey in~\cite{tosi2402nerfs} for a comprehensive overview.

Recent studies have also revisited classical Structure-from-Motion (SfM) techniques within a differentiable framework.
FlowMap~\cite{smith2024flowmap}, for example, presents an end-to-end model that jointly estimates camera intrinsics, extrinsics, and depth maps from monocular videos. 
Similarly, DUSt3R~\cite{wang2024dust3r} and MASt3R~\cite{leroy2024grounding} reformulate pairwise reconstruction as point map regression, simultaneously inferring depth, pose, intrinsics, and pixel correspondences. 
Extensions such as MASt3R++~\cite{duisterhof2024mast3r, murai2024mast3r} further improve dense multi-view stereo pipelines. 
Building on these ideas, recent works~\cite{wang20243d, elflein2025light3r, yang2025fast3r,cabon2025must3r,jang2025pow3r,liu2025regist3r} have demonstrated fast and accurate joint estimation of camera pose and depth across entire video sequences, signaling significant progress toward scalable 4D reconstruction systems.


\section{Level 2 -- 3D scene components}
\label{sec:level2}
While low-level cues provide the geometric and positional foundations necessary for understanding the scene’s layout, they are typically insufficient for capturing higher-level semantics and object-level structures.
Moving forward from this, level 2 methods focus on recovering the detailed representations of individual elements, such as objects, humans, and architectural structures, as well as their spatial arrangement within a scene. 
An overview of representative approaches in this category is shown in Fig.~\ref{fig:level2}. 
We specifically categorize the approaches into two types: (1) small-scale 3D object/scene reconstruction and (2) large-scale 3D scene reconstruction.
In the following subsection, we begin by reviewing the key 3D representations that underpin these methods.

\subsection{Scene representations}
In recent years, a variety of 3D scene representations have been developed and adopted for static surface reconstruction. In this subsection, we first highlight the most commonly used representations and explain how they are typically integrated into modern neural network architectures.

\noindent \textbf{Point cloud.}
Point clouds, composed of discrete 3D points often enriched with attributes like color and normals, are a fundamental representation for surface geometry. 
Beyond basic points, surfels, points with orientation and radius, offer a richer representation and support differentiable rendering~\cite{pfister2000surfels, yifan2019differentiable, kerbl20233d}. 
This enables optimization of point properties such as position, color, and size. 
Recent methods like Neural Point-based Rendering~\cite{aliev2020neural, dai2020neural}, SynSin~\cite{wiles2020synsin}, Pulsar~\cite{lassner2021pulsar, kopanas2021point}, and ADOP~\cite{ruckert2022adop} incorporate learnable features to better capture appearance and shape. 
Others, including FVS~\cite{riegler2020free}, SVS~\cite{riegler2021stable}, and FWD-Transformer~\cite{cao2022fwd}, further improve rendering by warping point-based features to novel views for color prediction, enhancing reconstruction quality.

\noindent \textbf{Meshes.} 
Meshes, formed by connecting vertices with edges and polygons (typically triangles or quads), are widely used for representing complex 3D shapes due to their flexibility and computational efficiency~\cite{botsch2010polygon, shirman1987local}. 
To utilize this kind of 3D representation, neural networks are generally designed to predict vertex positions~\cite{burov2021dynamic, thies2019deferred}, while textured appearances are commonly achieved using per-vertex colors or UV-mapped textures. 
To integrate meshes into 3D reconstruction pipelines, differentiable mesh rendering is essential. 
Techniques such as OpenDR~\cite{loper2014opendr}, Neural Mesh Renderer~\cite{kato2018neural}, Soft Rasterizer~\cite{liu2019soft}, and Paparazzi~\cite{liu2018paparazzi} support gradient-based learning, while general-purpose renderers like Mitsuba~\cite{nimier2019mitsuba} and Taichi~\cite{hu2019taichi} enable mesh-based differentiable rendering via automatic differentiation.

\noindent \textbf{Neural radiance field (NeRF).}
Neural Radiance Fields (NeRF)~\cite{mildenhall2020nerf} represent 3D scenes as continuous volumetric fields instead of discrete geometry like point clouds or meshes. 
By using a neural network (typically an MLP), NeRF maps a 3D point and viewing direction to color and density values.
Rendering is achieved through volumetric integration along camera rays, where sampled densities and colors are accumulated to produce the final pixel color using differentiable volume rendering~\cite{max1995optical}.
This implicit representation has enabled high-quality novel view synthesis and has seen widespread applications in editing~\cite{martin2021nerf}, inverse rendering~\cite{zhang2021physg}, camera pose estimation~\cite{lin2021barf}, and avatar reconstruction~\cite{weng2022humannerf}.

Extensions like NeuS~\cite{wang2021neus} and VolSDF~\cite{yariv2021volume} integrate signed distance functions (SDFs) into NeRF to provide sharper surface definitions. 
These methods convert signed distances into opacities using differentiable mappings and optimize the scene by minimizing photometric losses. 
NeRFs~\cite{mildenhall2020nerf, niemeyer2021giraffe, barron2021mip, barron2022mip, verbin2022ref, li2023dynibar} have become a versatile tool across a wide range of tasks in computer vision and graphics. 
They have been successfully applied to scene editing~\cite{martin2021nerf, zhi2021place}, camera pose optimization~\cite{lin2021barf, truong2023sparf}, inverse rendering~\cite{zhang2021physg, boss2021nerd}, generalization to unseen scenes~\cite{yu2021pixelnerf, wang2021ibrnet}, acceleration~\cite{reiser2021kilonerf}, and free-viewpoint video generation~\cite{du2021neural, li2022neural}. 
NeRFs also support avatar modeling tasks such as face and body reenactment~\cite{peng2021animatable, weng2022humannerf}. 
Beyond graphics, their adaptability has also extended to fields like robotics~\cite{zhou2023nerf}, medical imaging~\cite{ruckert2022neat}, and even astronomy~\cite{levis2022gravitationally}, highlighting the broad applicability of neural volumetric representations.

\noindent \textbf{3D Gaussian splatting (3DGS).}
3D Gaussian Splatting (3DGS)~\cite{kerbl20233d} offers an efficient alternative to NeRFs by directly optimizing a set of 3D Gaussians, each defined by a position $\mu$, opacity $\alpha$, anisotropic covariance $\Sigma \in \mathbf{R}^{3\times3}$, and spherical harmonic (SH) coefficients $\mathcal{SH}$ to model view-dependent color $\mathbf{c}$:
\begin{equation}
    \mathbf{G} = \{(\mu_i, \Sigma_i, \mathbf{c}_i, \alpha_i)\}_{i=1}^N,
\end{equation}
where $N$ is the number of 3D Gaussian primitives.
Note that Spherical Harmonic (SH) is used for controlling the color of each Gaussian to accurately capture the view-dependent appearance of the scene.
Unlike NeRFs that rely on MLPs, 3DGS represents scenes with explicit primitives, enabling high-resolution rendering with significantly faster training.

\subsection{Small-scale 3D object/scene reconstruction}
Given the significant advantages of 3D reconstruction from video sequences, early research efforts focused substantially on regressing a mesh as a unified representation for geometric surface reconstruction.
A primary task within this domain involves reconstructing the surface given a fixed, bounded video input. 
Pioneering methods typically relied on Structure-from-Motion (SfM)~\cite{schonberger2016structure, ozyecsil2017survey,iglhaut2019structure,lindenberger2021pixel} and Multi-View Stereo (MVS)~\cite{schoenberger2016mvs,zhang2020visibility,Yang_2020_CVPR,wang2020patchmatchnet,gu2020cascade} pipelines to predict dense depth maps, which were subsequently fused into surfaces using techniques like Poisson reconstruction~\cite{kazhdan2013screened} or Delaunay triangulation~\cite{lee1980two}. 
For instance,  COLMAP~\cite{schoenberger2016sfm, schoenberger2016mvs} employs pixel-level SIFT features for matching and depth prediction. 
MVSNet~\cite{yao2018mvsnet} leverages deep neural networks to extract latent features and aggregate depth predictions across multiple frames.
However, the depth maps generated by these approaches are frequently noisy, and both Poisson reconstruction~\cite{kazhdan2013screened} and Delaunay-based methods exhibit high sensitivity to noise within the underlying point clouds. 
This sensitivity often results in reconstructed surfaces of compromised quality.

Building upon these foundations and driven by the development of more efficient 3D representations, recent reconstruction methods derive surfaces directly from implicit volumetric representations, notably Neural Radiance Fields (NeRF)~\cite{mildenhall2021nerf} and 3D Gaussian Splatting (3DGS)~\cite{kerbl3Dgaussians}. 
These approaches are capable of yielding high-quality geometry and superior view synthesis. 
Specifically, NeRF-based surface reconstruction techniques, such as NeuS~\cite{wang2021neus}, VolSDF~\cite{yariv2021volume}, NeAT~\cite{meng2023neat}, and Neuralangelo~\cite{li2023neuralangelo}, jointly optimize a Signed Distance Function (SDF) field alongside the radiance field, subsequently extracting a mesh via the Marching Cubes algorithm~\cite{lorensen1998marching}. 
Later on, with the development of 3DGS, 2DGS~\cite{huang20242d}, GOF~\cite{yu2024gaussian}, and PGSR~\cite{chen2024pgsr}, propose to obtain a mesh by applying Truncated Signed Distance Function (TSDF) fusion to multi-view depth renderings.
SuGaR~\cite{guedon2024sugar} integrates Gaussians within an SDF field to provide high-quality reconstructions.
The field continues to advance rapidly, with several recent techniques further enhancing Gaussian Splatting for surface reconstruction by introducing diverse innovations: 
recovering high-frequency details at scale~\cite{chen20253d}; 
enforcing geometric consistency across multiple viewpoints~\cite{kim2025multiview}; 
utilizing triangulation constraints specifically for indoor scenes (Tri2plane~\cite{xie2025tri}); 
applying elongation splitting and assimilation strategies for improved accuracy (ESA-GS~\cite{chen2025esa}); 
inferring Unsigned Distance Functions (UDFs) (GaussianUDF~\cite{li2025gaussianudf}); 
introducing sorted opacity fields (SOF~\cite{radl2025sof}); 
leveraging photometric and reflectance priors~\cite{bruneau2025multi}; 
employing learned initialization strategies (QuickSplat~\cite{liu2025quicksplat}); 
and modifying Gaussian primitive representations~\cite{jiang2025geometry, shen2024solidgs}.

In contrast to these optimization-based methods, feed-forward approaches have emerged to directly predict feature volumes via an end-to-end network. 
From these predicted feature volumes, representations such as signed distance functions (SDF) coupled with radiance fields (\eg, SparseNeuS~\cite{long2022sparseneus}, GenS~\cite{peng2023gens}, C2F2NeuS~\cite{xu2023c2f2neus}, UFORecon~\cite{na2024uforecon}, SuRF~\cite{peng2024surface}, ReTR~\cite{liang2023retr}) or 2D Gaussian representations (LaRa~\cite{chen2024lara}) can be efficiently extracted. 
These methods demonstrate considerable potential for achieving generalized and real-time surface reconstruction by directly regressing both geometry and appearance in a single forward pass. 
However, the substantial memory requirements inherent in constructing and processing large feature volumes can impose practical limitations on reconstruction fidelity and scalability.

Reconstructing the environment from a first-person view is also an important application of spatial intelligence from video~\cite{furnari2023special}. 
Recently, with the advancements of world models and embodied AI, there have been a few attempts in static scene reconstruction from egocentric videos~\cite{plizzari2024outlook}. 
Specifically, SceneScript~\cite{avetisyan2024scenescript} utilizes a large language model to predict scene description codes from point clouds, which are generated from visual SLAM. 
EgoLifter~\cite{gu2024egolifter} and Photoreal Reconstruction~\cite{lv2025egosplats} leverage 3DGS to optimize photo-realistic static scenes directly from egocentric videos.
OSNOM~\cite{plizzari2024spatial} proposes to achieve scene reconstruction by first lifting 2D observations to 3D, and then mathcing them over various timesteps.

\subsection{Large-scale 3D scene reconstruction}

Modeling large-scale 3D scenes presents inherent challenges due to the complex, multi-scale nature of the required parameterization and the absence of predefined spatial boundaries. 
NeRF++~\cite{nerfpp2020} pioneered solutions for this domain by decomposing the radiance field into distinct bounded foreground and inverse sphere-based background components. 
This foundational approach enabled photorealistic rendering, extending far beyond the immediate camera frustum. 
Subsequently, Mip-NeRF360~\cite{mipnerf3602022} addressed critical issues of aliasing and scale imbalance through integrated cone sampling, a non-linear distortion field, and online distillation, achieving high-fidelity 360° reconstructions.
Building upon this, Zip-NeRF~\cite{zipnerf2023} effectively integrated Mip-NeRF360’s anti-aliasing capabilities with accelerated hash-grid representations, achieving comparable quality while training an order of magnitude faster;
\yk{CityGS~\cite{liu2024citygaussian, liu2024citygaussianv2} and OctreeGS~\cite{ren2024octree} proposed a novel divide-and-conquer training and Level-of-Detail (LoD) strategy to achieve efficiency large-scale training and rendering;
CityGS-X~\cite{gao2025citygs} further adopted a batch-level multi-task rendering process to achieve more efficient modeling;
LODGE~\cite{kulhanek2025lodge} recently introduced a hierarchical LoD representation, which iteratively selects optimal subsets of Gaussians based on the camera distance, to make real-time rendering feasible even on memory-constrained devices.}


These foundational advances enabled further scaling to vast environments. 
Partitioning strategies emerged as a key solution, with Block-NeRF~\cite{blocknerf2022} and Mega-NeRF~\cite{meganerf2022} decomposing scenes into independent local networks to support neighborhood and city-block navigation. 
City-NeRF~\cite{citynerf2022} extended this concept through progressive network and dataset expansion, seamlessly integrating satellite-to-street-level perspectives. 
Concurrently, F2-NeRF~\cite{f2nerf2023} accommodated arbitrary camera trajectories via perspective warping, while Gaussian splatting approaches like Scaffold-GS~\cite{scaffoldgs2023} achieved efficient view-adaptive rendering by anchoring sparse Gaussians on learned scaffolds.


Complementing these representation and partitioning frameworks, regularization techniques are also proposed to enhance the reconstruction robustness. 
MonoSDF~\cite{monosdf2022} strengthened outdoor geometry by integrating monocular depth and normal cues within a signed distance function (SDF) formulation. 
Mixture-of-experts systems like SCALAR-NeRF~\cite{scalarnerf2023} and Switch-NeRF~\cite{switchnerf2023} employ shared decoders or learned gating mechanisms to fuse predictions from multiple local models, establishing a critical pathway toward real-time, appearance-consistent reconstruction at city scales.

Another major advancement in surface reconstruction from static videos lies in the development of online, end-to-end systems that support real-time, interactive, and scalable applications. 
A notable early contribution is NeuralRecon~\cite{sun2021neuralrecon}, which introduced real-time reconstruction based on TSDF volumes using 3D convolutional GRUs~\cite{cho2014learning}. 
Building on this foundation, later works incorporated more advanced techniques: 
TransformerFusion~\cite{bozic2021transformerfusion} leveraged transformer architectures;
Flora~\cite{wang2023flora} improved feature aggregation; 
and VisFusion~\cite{gao2023visfusion} introduced visibility-aware fusion and ray-based sparsification.
To further reduce reliance on strict photometric consistency, several methods (SimpleRecon~\cite{sayed2022simplerecon}, DG-Recon~\cite{ju2023dg}, CVRecon~\cite{feng2023cvrecon}, and FineRecon~\cite{stier2023finerecon}) begin integrating geometric priors from foundation models, enabling better use of both global and local contextual information. 
Self-supervised techniques, like MonoSelfRecon~\cite{li2024monoselfrecon}, have also contributed to this trend. 
More recently, GeoRecon~\cite{bei2024georecon} and DetailRecon~\cite{chu2025detailrecon} have advanced the field further by improving global structural consistency and preserving fine-grained surface details.

\section{Level 3 -- 4D dynamic scenes}
\label{sec:level3}

\begin{figure}[t]
  \centering
  \includegraphics[width=\linewidth]{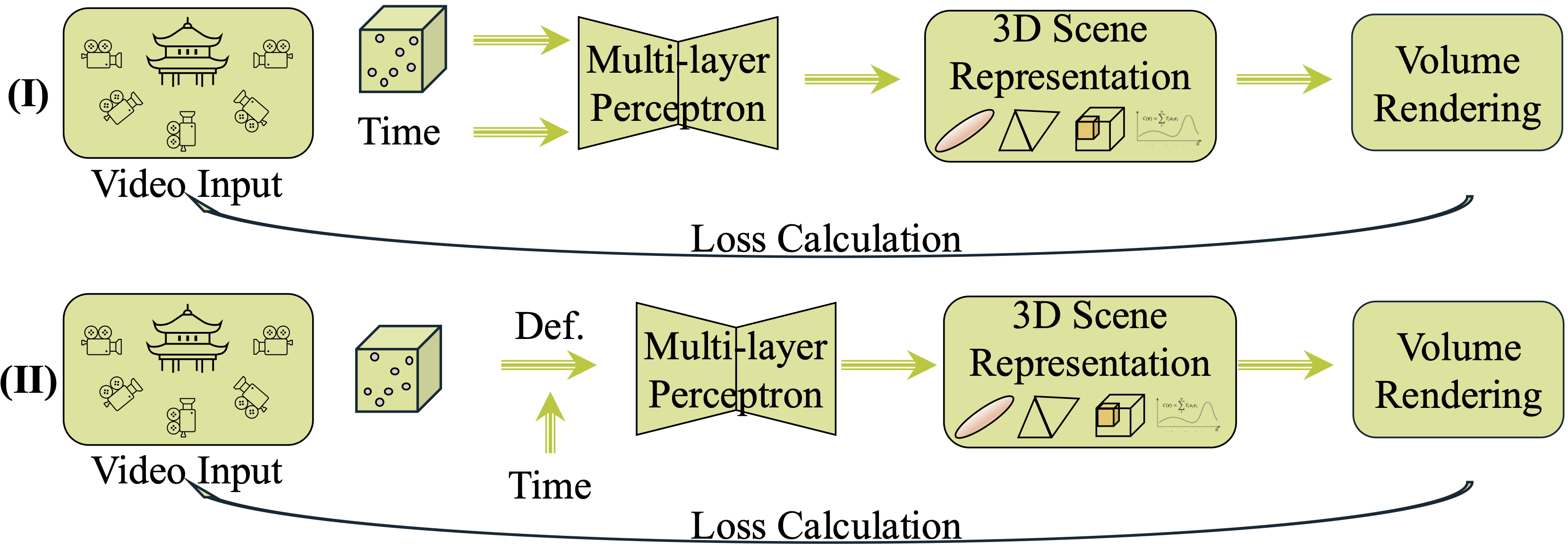}
  \caption{\textbf{The paradigms of methods for reconstructing dynamic scenes from video input.} Methods in this domain typically adopt one of two strategies for temporal modeling: (I) explicitly incorporating time as an additional input to extend a static 3D representation, or (II) reconstructing a canonical 3D space and learning its deformation over time. ``Def.'' denotes ``Deformation module''.}
  \label{fig:level3}
\end{figure}

The static nature of reconstructions from Level 2 methods limits their applicability in real-world, dynamic environments. To address this, Level 3 methods focus on introducing temporal dynamics into the scene, enabling the reconstruction of 4D representations that capture motion and changes over time. There are two popular directions (as illustrated in Fig.~\ref{fig:level3}), which are: (1) reconstruct a canonical space while learning its deformation over time, and (2) extend the original 3D representation by explicitly incorporating time as an additional input. Typically, these approaches can be broadly categorized into two groups based on their primary subjects: general 4D scene reconstruction and human-centric dynamic modeling. 

\subsection{General 4D scene reconstruction}

\noindent \textbf{Surface reconstruction from dynamic videos.}
Dynamic surface reconstruction from video is a vital area of research with wide-ranging applications in fields such as robotics, virtual reality, and autonomous systems. 
Early methods~\cite{casillas2021isowarp, kairanda2022f, zuffi20173d, li2020online} typically relied on deforming predefined object templates, which limited their ability to handle complex motions or occlusions. 
The advent of differentiable rendering~\cite{liu2019soft} has enabled more flexible reconstruction pipelines, allowing systems like LASR~\cite{yang2021lasr} and ViSER~\cite{yang2021viser} to model articulated shapes directly from video.
Building on this, several approaches have extended NeRF frameworks to dynamic and articulated objects, including BANMo~\cite{yang2022banmo}, PPR~\cite{yang2023ppr}, and REACTO~\cite{song2024reacto}. 
The use of neural implicit 3D representations~\cite{mildenhall2021nerf, park2019deepsdf} has further removed the need for template constraints, enabling fully unconstrained reconstruction of dynamic scenes~\cite{maoneural, shao2023tensor4d, johnson2023unbiased, choe2023spacetime}.
More recently, with the development of 3DGS,  methods such as~\cite{liu2024dynamic, ma2024reconstructing, cai2024dynasurfgs, li2024dgns, wang2024space, zheng2025gstar, chen2024adaptive} have incorporated 3DGS, significantly improving reconstruction speed, temporal coherence, and robustness to challenging motion.

\noindent \textbf{Novel view synthesis from dynamic videos.}
Beyond reconstructing dynamic 4D scenes, both NeRF and 3D Gaussian Splatting (3DGS) have been widely adopted for generating novel viewpoints.
This free-viewpoint video delivers immersive visual experiences while enabling the creation of freeze-frame (bullet time) effects~\cite{magnor2005video}. 
One popular direction is to reconstruct a static canonical radiance field and learn its deformation with time, as introduced by D-NeRF~\cite{pumarola2021d}. 
Building upon this idea, several NeRF-based approaches~\cite{park2021nerfies, tretschk2021non, park2021hypernerf, yu2023dylin, fan2024spectromotion} employ scene-specific optimization to model non-rigid motions, dynamic appearances, and complex specular effects. 
In parallel, recent methods~\cite{wu20244d,yang2024deformable,bae2024per,huang2024sc,lu2024dn,lu20243d,liu2024modgs, das2024neural, lei2024mosca} based on 3D Gaussian splatting leverage explicit point-based representations to directly encode dynamic geometry and appearance, offering improved computational efficiency and easier editing. 
Rather than relying solely on deformation fields, some approaches~\cite{wang2023flow, gao2024gaussianflow,zhu2024motiongs} model motion using vector fields derived from optical flow~\cite{beauchemin1995computation}, providing an alternative and interpretable way to describe temporal dynamics.

Another line of work extends radiance field representations by explicitly incorporating time as an additional input, enabling true 4D reconstruction. 
NeRF-based approaches~\cite{li2021neural, gao2021dynamic, you2024decoupling, wang2021neural, lin2023neural, lou2024darenerf, yang2023emernerf} treat time as a learnable parameter, allowing the model to capture temporal changes in geometry and appearance. 
Some approaches~\cite{du2021neural, li2021neural} use temporal flow to regularize training, while others~\cite{yoon2020novel} apply depth-based warping to synthesize novel views, even under inconsistent depth estimates. 
Additional efforts~\cite{xian2021space, wang2024shape, liu2023robust, cao2023hexplane, fridovich2023k} enhance temporal modeling by embedding explicit time-aware features, improving both coherence and efficiency in capturing scene dynamics.
On the other hand, 3DGS–based frameworks~\cite{li2024spacetime, yang2023real, duan20244d} incorporate timestamps as additional Gaussian attributes, enabling real-time and high-fidelity dynamic view synthesis through explicit point-based rendering. 
However, these methods can struggle with maintaining geometric consistency across time. 
To overcome this, more recent GS-based approaches~\cite{stearns2024dynamic, kocabas2024hugs, duisterhof2023md, luiten2023dynamic, sun20243dgstream, das2024neural, katsumata2023efficient} represent 4D scenes as temporally evolving trajectories of 3D Gaussians. This formulation improves temporal coherence, enhances tracking robustness, and provides more accurate reconstruction of dynamic geometry.

Feed-forward approaches to dynamic scene reconstruction have opened new possibilities for real-time 4D modeling. 
As the first approach, MonoNeRF~\cite{tian2023mononerf} pioneered this direction by introducing a generalizable dynamic radiance field that jointly encodes spatial and temporal features. 
Similarly, \cite{van2022revealing} tackles occlusion in dynamic settings without per-scene optimization. 
FlowIBR~\cite{busching2024flowibr} reduces optimization time by leveraging pre-training on large static datasets, while \cite{zhao2024pseudo} shows that strong temporal and geometric consistency can be achieved without tuning appearance for each scene. 
Most recently, \cite{liang2024feed} advances the paradigm by aggregating information from all context frames to reconstruct target frames directly, further improving efficiency and generalization. DAS3R \cite{xu2024das3r} combines 3D Gaussian Splatting with DUSt3R to reduce the optimization difficulty typically associated with 3D Gaussian Splatting, and achieves more accurate background reconstruction results.
LINO-UniPS~\cite{li2025light}, built upon VGGT, leverages learnable light register tokens to decouple illumination and normal features. It further introduces a global cross-image attention mechanism to enhance multi-view lighting representation and normal consistency.

\begin{figure}[t]
  \centering
  \includegraphics[width=\linewidth]{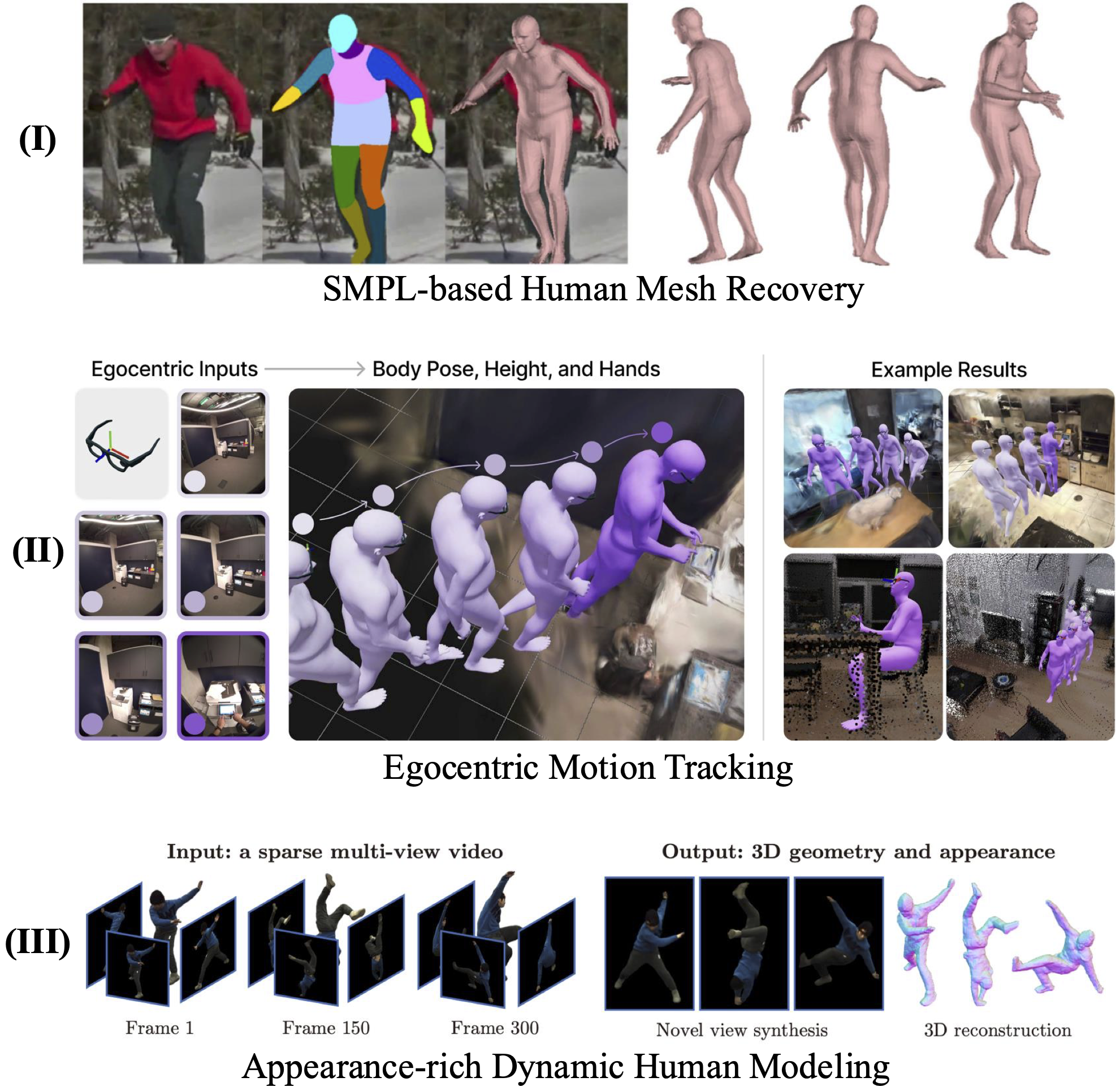}
  \caption{\textbf{The illustrations of methods for reconstructing 4D dynamic humans from video input.} Human-centric dynamic modeling approaches are generally categorized based on their representations: (I) methods that apply SMPL parametric model as their representation to derive the human pose and shape parameters (image source: Neural Body Fitting~\cite{omran2018neural}), (II) methods that similarly apply SMPL but focus more on the prediction based on egocentric videos (image source: EgoAllo~\cite{yi2025egoallo}), and (III) appearance-rich non-parametric methods that are capable of reconstructing the textured topologies, such as garments and accessories, from video data (image source: Neural Body~\cite{peng2021neuralbody}).}
  \label{fig:level3-human}
\end{figure}
\subsection{Human-centric dynamic modeling}
As illustrated in Fig.~\ref{fig:level3-human}, human-centric dynamic modeling methods can be grouped into two main categories based on their underlying 3D representation: SMPL-based human mesh recovery and appearance-rich dynamic human modeling. In the following, we will first illustrate SMPL~\cite{SMPL2015}, a parametric 3D human template, which forms the basis for human modeling.

\noindent \textbf{SMPL~\cite{SMPL2015}.} 
The parametric human model SMPL~\cite{SMPL2015} represents the 3D shape by incorporating body vertices, joints, face and hands landmarks, and expression parameters. Formally, given the pose parameter $\theta$ and shape parameter $\beta$, SMPL can map the canonical model with $n_{S}$ vertices to observation space:
\begin{equation} \label{smpl}
    \begin{aligned}
        M(\beta,\theta) &=\mathtt{lbs}(T(\beta,\theta),J(\beta), \theta,\mathcal{W}), \\ 
        T(\beta,\theta) &=\mathbf{T}+B_{s}(\beta)+B_{p}(\theta),
    \end{aligned}
\end{equation}
where $M$ is the function representing the SMPL model in the observation space, and $T$ gives the transformed vertices.
$\mathcal{W}$ is the blend weight, $B_s$ and $B_p$ are the shape blend shape function and pose blend shape function, respectively. $\mathtt{lbs}(\cdot)$ denotes the linear blend skinning function, corresponding to articulated deformation. It poses $T(\cdot)$ based on the pose parameters $\theta$ and joint locations $J(\beta)$, using the blend weights $\mathcal{W}$, individually for each body vertex:
\begin{equation}\label{eq:deformation}
    \mathbf{v}_o = \mathcal{G} \cdot \mathbf{v}_c, \quad \mathcal{G} = \sum_{k=1}^{K} w_k \mathcal{G}_k (\theta, {j}_k),
\end{equation}
where $\mathbf{v}_c$ and $\mathbf{v}_o$ respectively are SMPL vertices under the canonical pose and observation space, $w_k$ is the skinning weight, $\mathcal{G}_k (\theta, j_k)$ is the affine deformation that transforms the $k$-th joint ${j}_k$ from the canonical space to observation space, and $K$ is the number of neighboring joints.

SMPL-X~\cite{SMPL-X:2019} evolves from SMPL to include more face vertices, expression parameters $\phi$, and the expression blend shape function $B_e$ into the model:
\begin{equation} \label{smplx}
    \begin{split}
        M(\beta, \theta, \phi) &=\mathtt{lbs}(T(\beta, \theta, \phi), J(\beta), \theta, \mathcal{W}), \\
    T(\beta, \theta, \phi) &=\mathbf{T} + B_s(\beta) + B_e(\phi) + B_p(\theta).
    \end{split}
\end{equation}

\noindent \textbf{SMPL-based human mesh recovery and tracking}
Human mesh recovery (HMR) from dynamic videos has garnered significant research attention in recent years, facilitated by the use of parametric models like SMPL~\cite{SMPL2015}.
Early work approached the problem frame-by-frame using optimization~\cite{bogo2016keep,pavlakos2019expressive} or deep learning~\cite{kanazawa2018hmr,pavlakos2018learning} to estimate human pose and shape.
Progress in this area includes enhancements to camera modeling~\cite{SPEC:ICCV:2021}, graph-based or location-aware estimators~\cite{kolotouros2019convolutional,li2022cliff,sarandi2024nlf}, hybrid optimization and regression~\cite{kolotouros2019learning,song2020human}, kinematic parts and dense correspondences~\cite{omran2018neural,guler2019holopose,xu2019denserac,zhang2019danet,zeng20203d,georgakis2020hierarchical,kocabas2021pare}, image-aligned features~\cite{moon2020i2l,zhang2021pymaf,zhang2023pymafx}, and physical constraints~\cite{shimada2020physcap,tripathi2023moyo}.
Building on the success of transformer architectures~\cite{vaswani2017attention,dosovitskiy2020vit}, recent innovations emphasize tokenized representation~\cite{lin2021end,goel2023hmr2,dwivedi2024tokenhmr,saleem2024genhmr} and scaling up models for human pose and shape estimation to millions of training instances~\cite{cai2024smpler}. 
To capture motion over time, video-based HMR methods integrate temporal information using recurrent networks~\cite{kanazawa2019learning,kocabas2020vibe,wan2021encoder}, VAEs~\cite{rempe2021humor}, and optical flow~\cite{doersch2019sim2real}. 
Moreover, to tackle the challenges in estimating globally consistent human motion from unconditioned and dynamic cameras, researchers have adopted BERT-like~\cite{devlin2018bert} masked pretraining for motion sequence modeling~\cite{yuan2022glamr}, structure-from-motion (SfM)~\cite{ye2023decoupling}, optical flow~\cite{sun2023trace}, and simultaneous localization and mapping (SLAM)~\cite{shin2024wham,wang2024tram} techniques to holistically model human motion and camera movements in a shared world coordinate. 


\noindent\textbf{Egocentric motion tracking.}
Different from exocentric videos, egocentric videos are captured with head-mounted devices, \eg, smart glasses~\cite{engel2023projectaria} or VR/AR devices, from the first-person view, carrying rich information of the dynamic world, including wearers themselves. 
The camera motion in egocentric videos reflects the head movement and can be used as a conditional signal for full-body motion generation~\cite{li2023egoego,ma2024nymeria}, typically referred to as one-point body tracking.
Additional environmental cues, such as explicit hand detection~\cite{yi2025egoallo} or implicit video features~\cite{guzov2024hmd2,hong2025egolm, hatano2025invisible}, further enhance the performance of one-point tracking.
In VR applications, handheld controllers offer additional hand trajectories, providing extra constraints for ``three-point body tracking''. This would make the full-body motion tracking model grounded~\cite{castillo2023bodiffusion,du2023avatarsgrowleg,jiang2022avatarposer,hong2025egolm}.
However, both one-point and three-point tracking remain ill-posed problems due to the lack of lower-body observations. 
To mitigate this, wide field-of-view (FoV) cameras mounted on the head and angled downward are often employed to capture more of the body, improving the accuracy in full-body motion reconstruction~\cite{wang2023egocentric,wang2023scene}.


\noindent \yk{\textbf{Appearance-rich dynamic human modeling}}
Beyond HMR and interactive behavior capture, reconstructing animatable human avatars from RGB videos has emerged as a prominent research direction~\cite{wang2024survey}, encapsulating novel pose animation and novel view synthesis. 
Early methods relied on explicit geometric representations such as meshes, skeletons and silhouettes to model articulated motions and non-rigid surface deformations~\cite{xu2018monoperfcap}. 
A notable milestone was VideoAvatar~\cite{alldieck2018videoavatar}, which introduced canonical space mapping to decouple pose estimation from geometry and texture learning. 
Built upon this network, LiveCap~\cite{habermann2019livecap} further improves the efficiency to achieve real-time performance.
However, explicit representations faced limitations due to their dependency on pre-defined avatar templates, prompting a shift toward implicit neural representations. 
Pioneering approaches in this domain include regression models for avatar template prediction~\cite{guo2021human}, latent code aggregation across sequential observations~\cite{peng2021neuralbody}, and Fourier occupancy fields (FOF) for rapid reconstruction~\cite{feng2022fof}. 
Further developments extended NeRF to articulated human structures via generative models~\cite{su2021anerf}, motion field-based canonical warping~\cite{weng2022humannerf}, and hybrid frameworks integrating volumetric rendering with background reconstruction~\cite{jiang2022neuman,guo2023vid2avatar}. 
Additional strategies explored disentangled representations via graph neural networks~\cite{su2022danbo} and hybrid implicit-explicit representations for spatio-temporally coherent avatars~\cite{jiang2022selfrecon}. 
More recently, 3D Gaussian Splatting (3DGS)\cite{kerbl3Dgaussians} has enabled high-fidelity, animatable avatars with fast and flexible rendering\cite{hu2024gaussianavatar, kocabas2024hugs, li2024animatable, zheng2024gps, qian20243dgsavatar}, setting new standards in both reconstruction quality and view synthesis.

\section{Level 4 -- Interactions among scene components}
\label{sec:level4}

Advancing from previous levels, level 4 of spatial intelligence enters a more mature phase. At this level, the methods focus on modeling interactions between different components within a scene. 
Considering human is often the central subject of interaction, our following discussion emphasizes human-centric interaction modeling.
These approaches are also categorized into three groups based on their input types and the underlying representations: SMPL-based human-centric interaction, appearance-rich human-centric interaction, and egocentric human-centric interaction.

\begin{figure}[t]
  \centering
  \includegraphics[width=\linewidth]{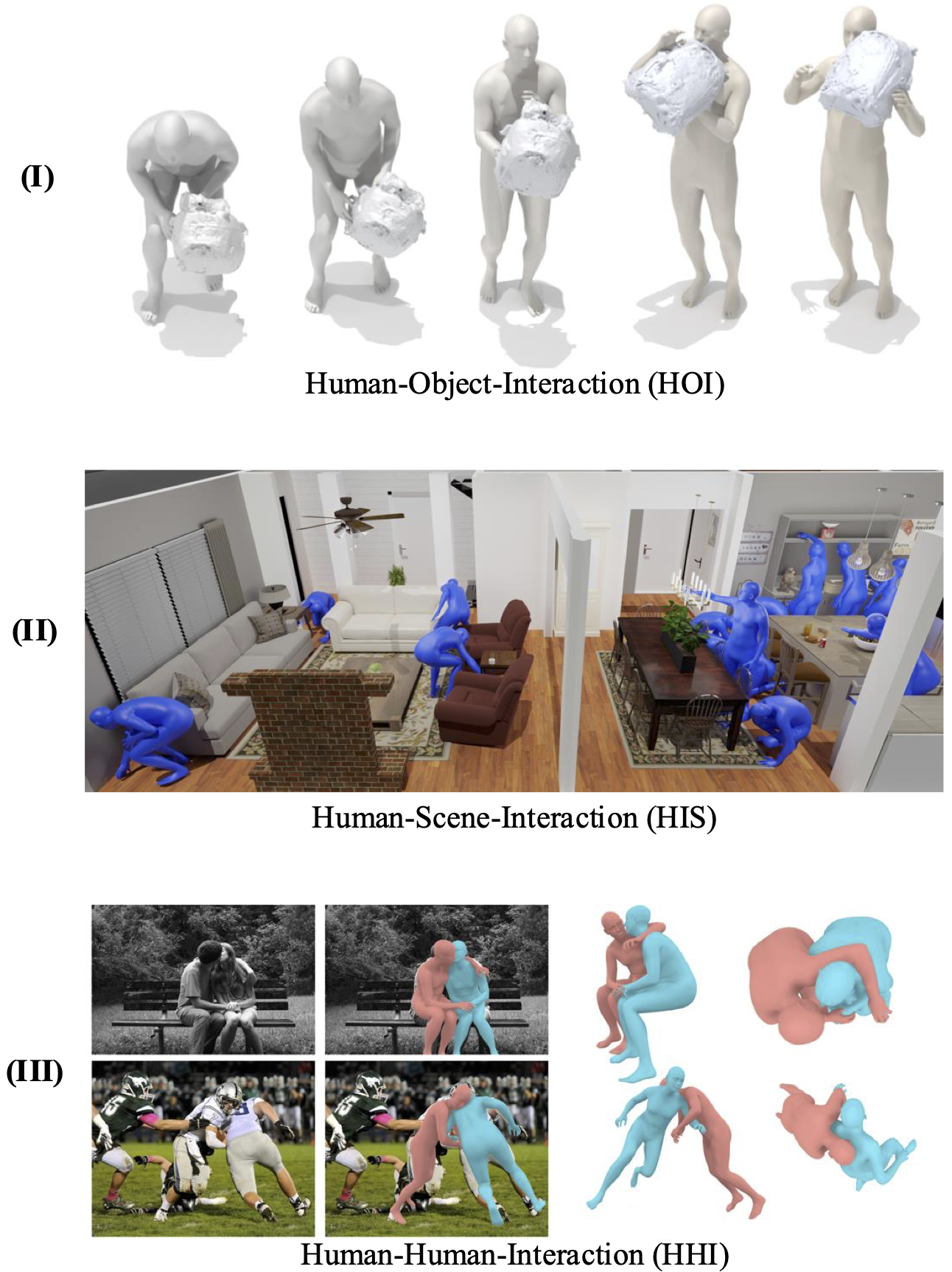}
  \caption{\textbf{Examples of methods for modeling SMPL-based human-centric interaction.} Image source: InterDreamer~\cite{xu2024interdreamer}, CIRCLE~\cite{araujo2023circle}, and BUDDI~\cite{muller2024generative}.}
  \label{fig:level4-smpl}
\end{figure}
\subsection{SMPL-based human-centric interaction}
Building on human mesh recovery (HMR), recent research has progressed toward capturing 3D interactive behaviors from videos, which can be broadly classified into three categories: human-object interaction (HOI), human-scene interaction (HSI), and human-human interaction (HHI). Examples of different categories is provided in Fig.~\ref{fig:level4-smpl}

\noindent \textbf{Human-object-interaction (HOI).}
Considering the lack of high-quality 3D HOI data with videos~\cite{savva2016pigraphs}, early methods tackled 3D HOI by utilizing traditional optimization frameworks to reconstruct human-object spatial arrangements through heuristic contact priors~\cite{zhang2020phosa, goel2023humans}. 

The emergence of scalable methods for collecting 3D HOI data involving videos~\cite{dabral2021gravity,xu2021d3d,bhatnagar2022behave,huang2022intercap,jiang2023full, ma2024nymeria} prompted learning-based approaches, which present significant advantages in generalization and inference efficiency. Pioneering works proposed to model object proximity relative to the human by learning signed distance fields (SDFs) from data~\cite{xie2022chore,xie2023visibility}, and then operate post-optimization based on the learned field. To enhance robustness and accuracy, particularly under occlusions, methods based on generative models such as normalizing flows~\cite{huo2024stackflow,huo2024monocular} learn the distribution of human-object spatial arrangements conditioned on input video, thereby mitigating outlier predictions. A separate strategy augments auxiliary input modality, such as IMU, to facilitate object tracking~\cite{zhao2024imhoi}. 

However, a persistent limitation of these methods is their reliance on well-reconstructed object template geometries, restricting their feasibility and applicability across diverse scenarios. To overcome this challenge, recent approaches like HDM~\cite{xie2024template} and InterTrack~\cite{xie2024intertrack} proposed to learn geometric correspondences within families of similar object categories using diffusion models~\cite{ho2020denoising,song2020denoising}. This enabled geometry-agnostic reconstruction of 3D HOI point clouds directly from image frames and facilitated the construction of large-scale synthetic 3D HOI datasets.
 
\noindent \textbf{Human-scene-interaction (HSI).}
Extending to full scenes including movable objects and fixed contextual layout, early methods focused on estimating contacts between human and static scenes from image frames~\cite{hassan2019resolving,huang2022capturing}, which are trained on data with sparse labels and inaccurate 3D scene geometries.

To address the limitations of scale and quality of comprehensive 3D scene modeling in HSI data, GTA-IM~\cite{cao2020long} constructed synthetic data comprising videos alongside pseudo 3D HSI labels obtained from 3D assets within the game engine. Similarly, CIRCLES~\cite{araujo2023circle} integrated real-world motion capture with digital environments via VR applications, while TRUMANS~\cite{jiang2024scaling} replicated 3D scene assets in reality. These methods provided richer and more accurate 3D labels, enabling reconstruction of HSI from videos to advance into diverse indoor and outdoor contexts involving dynamic objects.

Nevertheless, a significant gap persists between 3D assets and real-world environments. Jointly reconstructing both humans and dynamic scenes from casual, real-world videos like web footage remains highly desirable. SitComs3D~\cite{pavlakos2022one} targeted on television show with multiple shots of the same scene. By disentangling human and scene using different representations, SitComs3D~\cite{pavlakos2022one} expressed the scene as NeRF~\cite{mildenhall2020nerf} and estimated human motion within this context.  More recently, leveraging advanced low-level 3D attributes prediction models (introduced in Level 1), JOSH~\cite{liu2025josh} jointly recovered human motion, 3D scene structure, camera poses, contacts, and optimized them contextually with physics-based constraints. While ODHSR~\cite{zhang2025odhsr} achieved the same target by holistically representing the human and scene as 3DGS for each frame.

\noindent \textbf{Human-human-interaction (HHI).}
In the case of multi-person interactions, earlier monocular and sparse multi-view systems utilized 3D keypoint heatmaps for multi-human pose estimation~\cite{sun2021romp,shuai2023closecap}.
However, these methods ignore geometric constraints and physical contacts, leading to unrealistic results.

To address this issue, datasets~\cite{yin2023hi4d,khirodkar2024harmony4d} and methods~\cite{lu2024avatarpose} introduced instance-level prior and geometric collision loss to obtain physically plausible multi-human interactions from multi-view videos. While BUDDI~\cite{muller2024generative} and HumanInteraction~\cite{huang2024closely} leveraged generative models such as diffusion models~\cite{ho2020denoising,song2020denoising} and VQ-VAE~\cite{van2017vqvae} to model interaction prior, which effectively provides a desirable initial estimation for following optimization iterations. MultiPhys~\cite{ugrinovic2024multiphys} take another strategy to incorporate a physical simulator to search for the optimal policy in physically correct motion space via an imitation learning framework.


\yk{\subsection{Appearance-rich human-centric interaction}}
Earlier attempts at reconstructing textured human-centric interactions from videos were limited by the lack of high-quality 3D datasets, making it difficult to model complex interactions between humans and objects. 
Fortunately, recent progress in differentiable 3D representations, particularly NeRF and 3D Gaussian Splatting (3DGS), has opened up new possibilities for capturing these interactions without relying on expensive 4D datasets.

A representative example is HOSNeRF~\cite{liu2023hosnerf}, which enables joint reconstruction of humans and their interacted objects (e.g., backpacks) from RGB videos. Specifically, it extends the human skeleton with object bones, allowing the model to account for deformations introduced by contact. The process to obtain color $\mathbf{c}$ and density value $\sigma$ for each point $\mathbf{x}_c$ can be then written as:
\begin{equation}
    F(\gamma(\mathbf{x}_c), \mathcal{O}_c^i) \mapsto (\mathbf{c}, \sigma),
\end{equation}
where $F(\cdot)$ is the NeRF module, $\gamma(\cdot)$ denotes the positional embedding, and $\mathcal{O}_c^i$ is the learnable state embedding representing object states in the canonical space at frame $i$. These embeddings allow the model to conditionally represent different interaction configurations.

Following HOSNeRF, other recent methods further extend this direction: NeuMan~\cite{jiang2022neuman} decouples the human and scene by training separate NeRFs for each, improving flexibility and scene composition;
PPR~\cite{yang2023ppr} combines differentiable physics simulation with differentiable rendering, optimizing the reconstruction via coordinate descent to improve realism;
RAC~\cite{yang2023reconstructing} generalizes the approach to animals and humans by learning consistent skeletons with fixed bone lengths.

\begin{figure}[t]
  \centering
  \includegraphics[width=\linewidth]{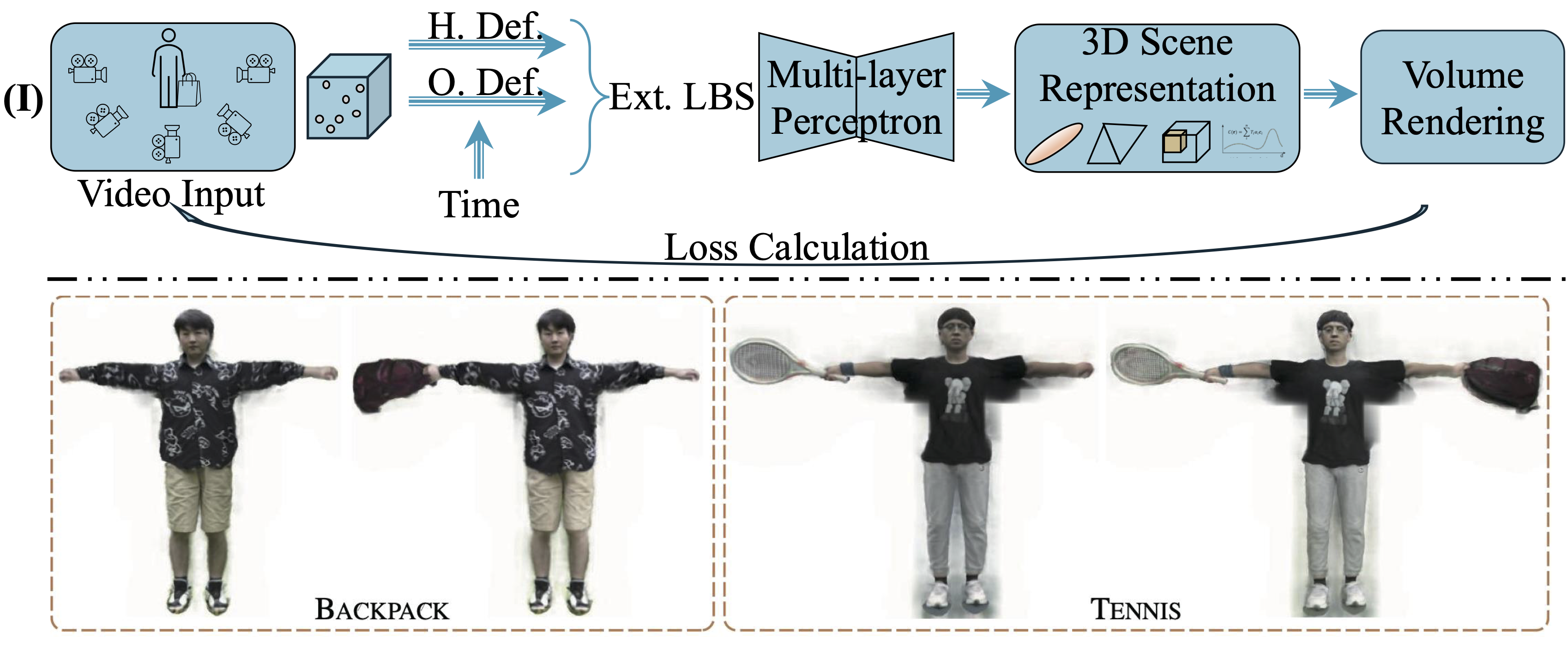}
  \caption{\textbf{The paradigms of methods for reconstructing appearance-rich human-centric interaction.} These methods generally build on SMPL-based linear blend skinning (LBS) deformation, extending the human body skeleton to include interacted objects. An example result is shown in the figure below. (image source: HOSNeRF~\cite{liu2023hosnerf}). ``H. Def.'', ``O. Def.'', and ``Ext. LBS'' denote ``Human Deformation'', ``Object Deformation'', and ``Extended SMPL-based LBS'' correspondingly.}
  \label{fig:level4-textured-human}
\end{figure}
\subsection{Egocentric human-centric interaction}
Egocentric videos, captured from a first-person perspective, uniquely record the wearer's interactions with objects, environments, and other people, offering rich context for reconstructing and understanding human-centric interactions.
Most existing benchmarks and models primarily focus on hand-object interactions~\cite{damen2020human}.
An early effort in this domain, H2O~\cite{kwon2021h2o}, captured egocentric hand-object interactions using a head-mounted RGB-D camera along with multiple third-person cameras.
HOI4D~\cite{liu2022hoi4d} further scales up the egocentric hand-object interaction capture with more objects. 
HOT3D~\cite{banerjee2024introducinghot3d} leverages Project Aria glasses~\cite{engel2023projectaria} and Quest 3 headsets in a multi-camera rig to enable more precise annotation of hand and object poses.
Beyond household scenarios, egocentric hand-object interaction also plays a crucial role in other domains. 
For instance, POV-Surgery~\cite{wang2023pov} introduces a synthetic dataset tailored for estimating hand and instrument poses in surgical settings; 
HOI-Ref~\cite{bansal2024hoi} curates an HOI-QA dataset that consists of 3.9M question-answer pairs and achieves good performance for retrieving human-object-interactions from videos;
AMEGO~\cite{goletto2024amego} introduces a new Active Memories Benchmark (AMB) while achieving SOTA performance for capturing key locations and object interactions;
Ego-Exo4D~\cite{grauman2024ego}, Nymeria~\cite{ma2024nymeria}, HD-EPIC~\cite{perrett2025hd}, and EPIC-Fusion~\cite{darkhalil2022epic, kazakos2019epic} collect the first large-scale 4D datasets for human scene interactions.
More recently, works like EPIC-Fields~\cite{tschernezki2023epic} present the first trial to leverage the 3D priors for understanding the videos with human-centric interactions.

\section{Level 5 -- Incorporation of physical laws and constraints}
\label{sec:level5}

\begin{figure}[t]
  \centering
  \includegraphics[width=\linewidth]{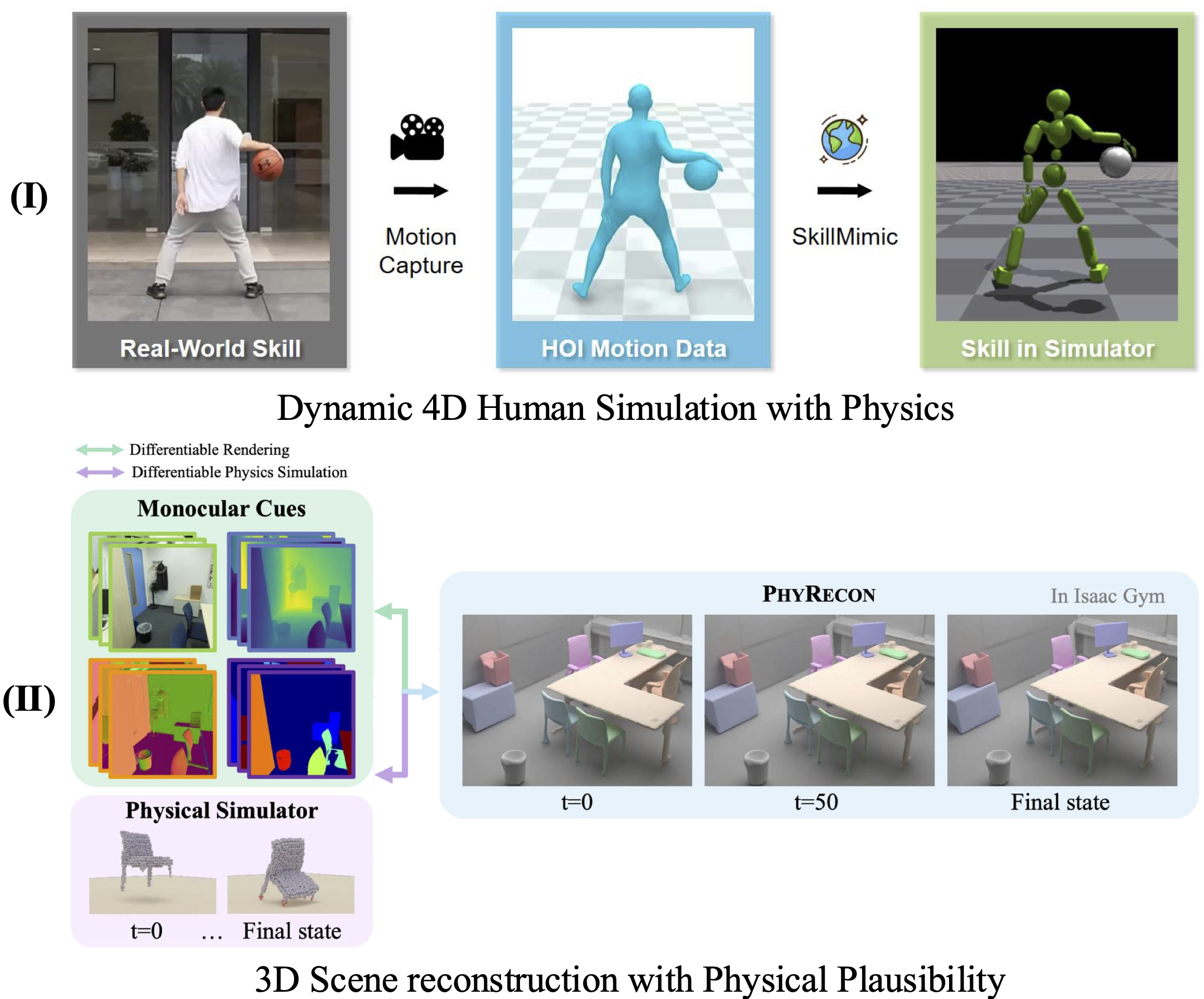}
  \caption{\textbf{The paradigms of methods for inferring physically grounded 3D spatial understanding from videos.} (I) Physical dynamic human modeling methods learn motion policies from real-world captures of human-object interactions, enabling deployment in simulators and transfer to humanoid robotics (image source: SkillMimic~\cite{wang2024skillmimic}). (II) Physically plausible 3D scene reconstruction mitigates missing geometry artifacts prevalent in traditional approaches, producing simulation-ready environments (image source: PhyRecon~\cite{ni2024phyrecon}).}
  \label{fig:level5}
\end{figure}

Recent advancements in 4D scene reconstruction and embodied AI have paved the way for a shift in research focus, \ie, modeling the underlying physics of the environment to achieve a more comprehensive representation of 4D spatial intelligence that imitates real-world human experiences. This evolution in research aims not only to capture dynamic human-scene interactions over time but also to embed physical plausibility and reasoning into these reconstructions. Such developments are crucial for enabling intelligent robotic systems to operate effectively in complex, unstructured environments. As illustrated in Fig.~\ref{fig:level5}, current efforts in this area primarily focus on 4D reconstruction of dynamic humans and 3D scenes. While some recent methods~\cite{li2025dso} have incorporated physics to improve 3D object generation, we do not cover this line of work in our illustration.

\subsection{Dynamic 4D human simulation with physics}
Recent advancements in physical human modeling have focused on generating realistic, physics-based character animation using motion capture data, imitation learning, and reinforcement learning. This body of work spans two key areas: physics-driven character animation and human-object interaction (HOI).

\noindent \textbf{Physics-based character animation}
Generating physically accurate motion for human and animal characters has long been a central challenge in animation and control research~\cite{lee2002interactive, lee2010motion, safonova2007construction, treuille2007near, levine2012continuous, zhang2018mode, holden2017phase, ling2020character}. Recent advances in physics-based character animation emphasize learning from large MoCap datasets using reinforcement learning (RL)~\cite{messikommer2024reinforcement} and imitation learning~\cite{2018-TOG-deepMimic, wang2020unicon, won2020scalable, Luo2023PerpetualHC, ScaDiver, wagener2022mocapact}. 
Among them, a foundational approach is DeepMimic~\cite{2018-TOG-deepMimic}, which learns to reproduce dynamic motions through direct trajectory tracking. AMP~\cite{peng2021amp}, based on GAIL~\cite{ho2016generative}, improves realism by using a generative adversarial framework, where a discriminator judges the realism of the motion, guiding the controller during training. However, AMP demands training a separate policy for each task. Extensions like ASE~\cite{peng2022ase}, CALM~\cite{tessler2023calm}, ControlVAE~\cite{ControlVAE}, PULSE~\cite{luo2024universal}, OmniGrasp~\cite{luo2024omnigrasp}, HOVER~\cite{he2024hover}, ASAP~\cite{he2025asap}, and UniPhys~\cite{wu2025uniphys} aim to extract more general motion priors that can be reused across tasks. MaskedMimic~\cite{tessler2024masked} introduces a masked conditional VAE for multi-task learning, but still struggles with generalizing to unseen control signals. 

In parallel, there has been a rise in text-driven control approaches~\cite{tessler2023calm, juravsky2022padl, juravsky2024superpadl, truong2024pdp}, where high-level natural language is used to direct character behavior. For instance, SuperPADL~\cite{juravsky2024superpadl} employs a multi-stage training pipeline combining reinforcement learning and behavior cloning. PDP~\cite{truong2024pdp} uses diffusion models to create multimodal controllers that interpret text commands, improving robustness by injecting noise during training. 
Despite these innovations, text-driven physical controllers still lag behind kinematic methods in expressiveness and diversity due to difficulties in distilling controllable and reliable multimodal behaviors.
To bridge this gap, researchers have turned to hierarchical control frameworks~\cite{tevet2024closd, hansen2024hierarchical, rempeluo2023tracepace, Wang2024PacerPlus}. These methods split the problem into a high-level planning stage and a low-level controller. The planner might generate trajectories~\cite{tevet2024closd}, waypoints~\cite{rempeluo2023tracepace}, or partial-body targets~\cite{hansen2024hierarchical}, which are tracked by an RL policy. For example, CLoSD~\cite{tevet2024closd} combines a kinematic diffusion-based planner with a physics-based tracker. However, mismatches between high-level kinematic plans and low-level physical feasibility often lead to artifacts like foot sliding or jitter, which require task-specific fine-tuning~\cite{tevet2024closd, hansen2024hierarchical}.

\noindent \textbf{Learning human-object interaction (HOI)}
Human-object interaction presents additional complexity due to the need for fine contact control, multi-body coordination, and realistic physical responses. Early systems approached HOI using handcrafted state machines or models like inverted pendulums to simulate behaviors such as running or jumping~\cite{hodgins1995animating, yin2007simbicon, coros2010generalized}. More recent works use deep RL~\cite{arulkumaran2017deep} to model more diverse interactions, including sports and tool use~\cite{liu2018learning, liu2017learning, tan2014learning}.
Imitation learning is a natural fit for HOI, and methods like~\cite{tennis, bae2023pmp, braun2023physically} attempt to model whole-body interactions using motion priors or grasping models. However, directly adapting locomotion-based imitation frameworks (\eg, DeepMimic~\cite{2018-TOG-deepMimic}, AMP~\cite{peng2021amp}) to HOI has proven unstable. These methods often fail to address unbalanced reward structures, crucial contact dynamics, or the importance of relative positioning, leading to poor performance.
Some approaches use interaction graphs to model spatial dependencies between body and object~\cite{ho2010spatial, zhang2023simulation}, but these primarily focus on kinematics and often lack physical realism. In contrast, more recent frameworks introduce contact-aware rewards that explicitly encourage correct and stable contact, significantly improving performance on complex HOI tasks. This contact-driven perspective, introduced in early work like~\cite{wang2023physhoi, wang2024skillmimic}, allows for unified training across a wide range of interaction scenarios without the need for handcrafted rewards or separate pipelines.

\subsection{3D scene reconstruction with physical plausibility}
As we move toward building the 3D virtual worlds that can mimic real-world actions, physically plausible 3D scene reconstruction is gradually becoming a central focus in 3D scene modeling~\cite{zhang2024physdreamer, ni2024phyrecon}.
Addressing this challenge, PhysicsNeRF~\cite{barhdadi2025physicsnerf} injects explicit physics guidance--specifically, depth-ranking, sparsity, and cross-view alignment losses--to achieve stable and physically consistent geometry even from extremely sparse multi-view inputs. Building on this foundation, inverse-rendering pipelines such as PBR-NeRF~\cite{wu2025pbrnerf} integrate neural radiance fields with physics-based rendering priors. This coupling enables the joint optimization of geometry, illumination, and spatially varying materials, effectively mitigating the physically impossible albedo–lighting entanglement inherent in vanilla NeRFs. Progressing to the scene level, CAST~\cite{yao2025cast} first retrieves CAD proxies from a single RGB image and subsequently applies a physics-aware correction step. This step rigorously enforces support, non-penetration, and object-relation constraints, resulting in contact-consistent layouts. PhyRecon~\cite{ni2024phyrecon} proposes to leverage the differentiable gradients from the simulator~\cite{hu2019difftaichi, hu2019taichi} to improve the physical plausibility of the reconstruction scene components. Orthogonal to explicit simulators, Aug-NeRF~\cite{chen2022augnerf} employs triple-level, physically grounded augmentations as a regularization strategy during training. This technique dramatically reduces view-inconsistent floaters and enhances generalization capabilities. Finally, specialized methodologies are emerging for targeted phenomena; for instance, the Planar Reflection-Aware NeRF~\cite{gao2024reflectionnerf} explicitly models secondary reflected rays. This advancement eliminates the floaters frequently hallucinated behind reflective surfaces like glass and mirrors, thereby further improving the physical plausibility of reconstructions in everyday indoor scenes.



\section{Challenges and future directions}
\label{sec:challenges}

While notable advancements have been made for methods from level 1 to level 5, current techniques still encounter major challenges. 

\subsection{Level 1 -- low-level 3D attributes}
\noindent \textbf{Challenges}
Despite remarkable advances, reconstructing depth, camera, and 3D tracking from video remains challenging, due to its inherent ill-posed nature, especially for unconstrained, dynamic inputs. 
\textbf{(1)} A core challenge lies in handling occlusions, dynamic object motion, and non-Lambertian surfaces, which violate many of the assumptions underlying existing methods.
\textbf{(2)} Many methods still require post-processing steps, global alignment, or manual hyperparameter tuning, limiting their automation and applicability.
\textbf{(3)} Ensuring robustness to sensor noise, and generalization to diverse viewpoints and motion patterns (\eg, handheld, drone, egocentric) still remain open concerns.

\noindent \textbf{Future directions}
Given the key challenges outlined above, several promising directions can be explored.
\textbf{(1)} Developing world models that jointly represent geometry, motion, semantics, and uncertainty could offer a principled approach to reducing ambiguity in 4D reconstruction. These models can also take cues from vision-language foundation models, using large-scale pretraining on both synthetic and real-world video to learn strong inductive biases.
\textbf{(2)} Additionally, further progress can also be made to achieve interactive annotation tools and multi-agent data collection, to provide richer supervision for training more robust systems.

\subsection{Level 2 -- 3D scene components}
\noindent \textbf{Challenges}
While 3D scene reconstruction from video has seen remarkable progress, multiple technical challenges remain.
\textbf{(1)} There is no universally optimal scene representation. Point clouds, meshes, NeRFs, and 3D Gaussian Splatting each present trade-offs between fidelity, efficiency, and expressiveness.
\textbf{(2)} Recovering fine-scale geometry in unbounded or textureless regions, particularly under challenging conditions such as motion blur, lighting changes, or sparse viewpoints, remains difficult.
\textbf{(3)} Reconstructions from egocentric videos are especially prone to degradation due to rapid motion and limited field of view (FoV), often resulting in incomplete or distorted outputs.

\noindent \textbf{Future directions}
In light of these challenges, future directions could potentially focus on:
\textbf{(1)} Developing hierarchical and scalable architectures, such as mixture-of-experts models, sparse voxel grids, and scaffolded splatting, might be key to enabling efficient reconstruction and rendering across large-scale environments.
\textbf{(2)} Advancing egocentric and dynamic scene understanding through cross-modal learning, by integrating signals such as IMU data, audio, and textual descriptions, may improve robustness in complex and motion-heavy scenarios.

\subsection{Level 3 -- 4D dynamic scenes}
\noindent \textbf{Challenges}
Despite rapid progress in 4D scene reconstruction and human-centric dynamic modeling, several critical challenges remain unresolved:
\textbf{(1)} Feed-forward methods accelerate reconstruction but suffer speed-generalization-quality trade-offs, often relying on per-scene optimization or massive training data that hinder scalability.
\textbf{(2)} Complex dynamic phenomena, including fluids, smoke, semi-rigid objects, and topological changes (\eg, object splitting/merging), remain largely unsolved.
\textbf{(3)} Similarly, egocentric reconstruction is severely challenged by self-occlusions and limited FoV in head-mounted fisheye captures, complicating dynamic scene recovery.


\noindent \textbf{Future directions}                                                                                          
Based on these issues, several directions deserve to be considered.
\textbf{(1)} Hybrid implicit-explicit representations could balance reconstruction speed and fidelity, while physics-informed priors (biomechanics, fluid dynamics) may enforce physically plausible motion.
\textbf{(2)} Benchmarks and collaborative exo-ego capture frameworks are critical to evaluate temporal coherence, overcome occlusion limitations, and enable scalable AR/VR/robotics applications.   

\subsection{Level 4 -- interactions among scene components}
\noindent \textbf{Challenges}
Reconstructing human-centric interactions from video remains a highly challenging task due to several technical limitations. 
Human-object interaction (HOI) methods often require accurate object templates, which restricts their ability to generalize across diverse object categories and deformable instances.
Human-scene interaction (HSI) models struggle to consistently align humans with dynamic environments over time, particularly in real-world videos where spatial and temporal cues are sparse.
For human-human interaction (HHI), monocular setups frequently face issues such as occlusion, depth ambiguity, and unrealistic contact modeling.
Across all domains, maintaining physical plausibility, preventing interpenetration, and ensuring temporal coherence remain persistent challenges.
Moreover, the lack of large-scale, high-quality datasets that capture textured interactions or egocentric viewpoints continues to impede the method's generalization and real-world deployment.

\noindent \textbf{Future directions}
To address these challenges, future research could focus on category-agnostic modeling through geometry-aware or generative approaches such as diffusion models, to enable broader generalization across interaction types. 
Integrating differentiable physics, learned contact priors, or imitation learning frameworks may improve physical realism in interactions. 
For egocentric interactions, future progress may be able to come from multimodal fusion (e.g., video, IMU, gaze) and learning from large-scale benchmarks. 
Last but not least, achieving real-time and interactive simulation of human-centric behaviors might require combining video-based reconstruction with embodied reasoning and action modeling.

\subsection{Level 5 -- incorporation of physical laws and constraints}
\noindent \textbf{Challenges}
Despite significant progress, physics-based reconstruction and simulation from video still face some challenges. 
Reinforcement learning based character animation often suffers from sample inefficiency, high computational cost, and unstable optimization, particularly for complex or contact-rich motions.
Generalization across diverse motion types remains limited; most policies are task-specific, and hierarchical controllers frequently introduce artifacts due to mismatches between high-level kinematic planning and low-level physical feasibility.
For HOI situations, achieving stable contact behaviors and precise coordination between human and object dynamics remains difficult, especially under limited supervision or with diverse object geometries.
In 3D scene reconstruction, enforcing physical plausibility, such as support, friction, and non-penetration, is difficult when scene geometry is incomplete or inferred from sparse views. As a result, many methods continue to produce floating artifacts, interpenetrations, or physically implausible contacts due to insufficient physical priors.

\noindent \textbf{Future directions}
To overcome these limitations, future research could first explore multimodal representations that integrate video cues, physical constraints, and motion priors from different tasks and domains.
Secondly, differentiable physics engines may offer a promising foundation for jointly optimizing geometry, dynamics, and interaction constraints within a fully learnable pipeline.
Thirdly, developing hierarchical and diffusion-based motion planning, particularly when conditioned on text or scene context, may also lead to more expressive, controllable, and physically plausible character behaviors.
Finally, better integration of perception and control, allowing characters to adapt their behavior based on inferred scene dynamics, may unlock more interactive and physically realistic virtual humans.

\section{Conclusion}
In this report, we begin by categorizing 4D spatial intelligence into five distinct levels: Level 1 — basic 3D cues; Level 2 — components of 3D scenes; Level 3 — dynamic 4D scenes; Level 4 — interactions among scene components; and Level 5 — integration of physical laws and constraints.
We then provide a thorough review of methods corresponding to each level. Additionally, we discuss the remaining challenges faced by current techniques and explore promising future directions to overcome these issues.
As this is a rapidly evolving field with new papers published weekly or even daily, we hope this survey offers an accessible entry point for interested readers and inspires progress toward a potential Level 6 in 4D spatial intelligence.

\bibliographystyle{IEEEtran}
\bibliography{ref}

\begin{IEEEbiography}
[{\includegraphics[width=1in,height=1.25in,clip,keepaspectratio]{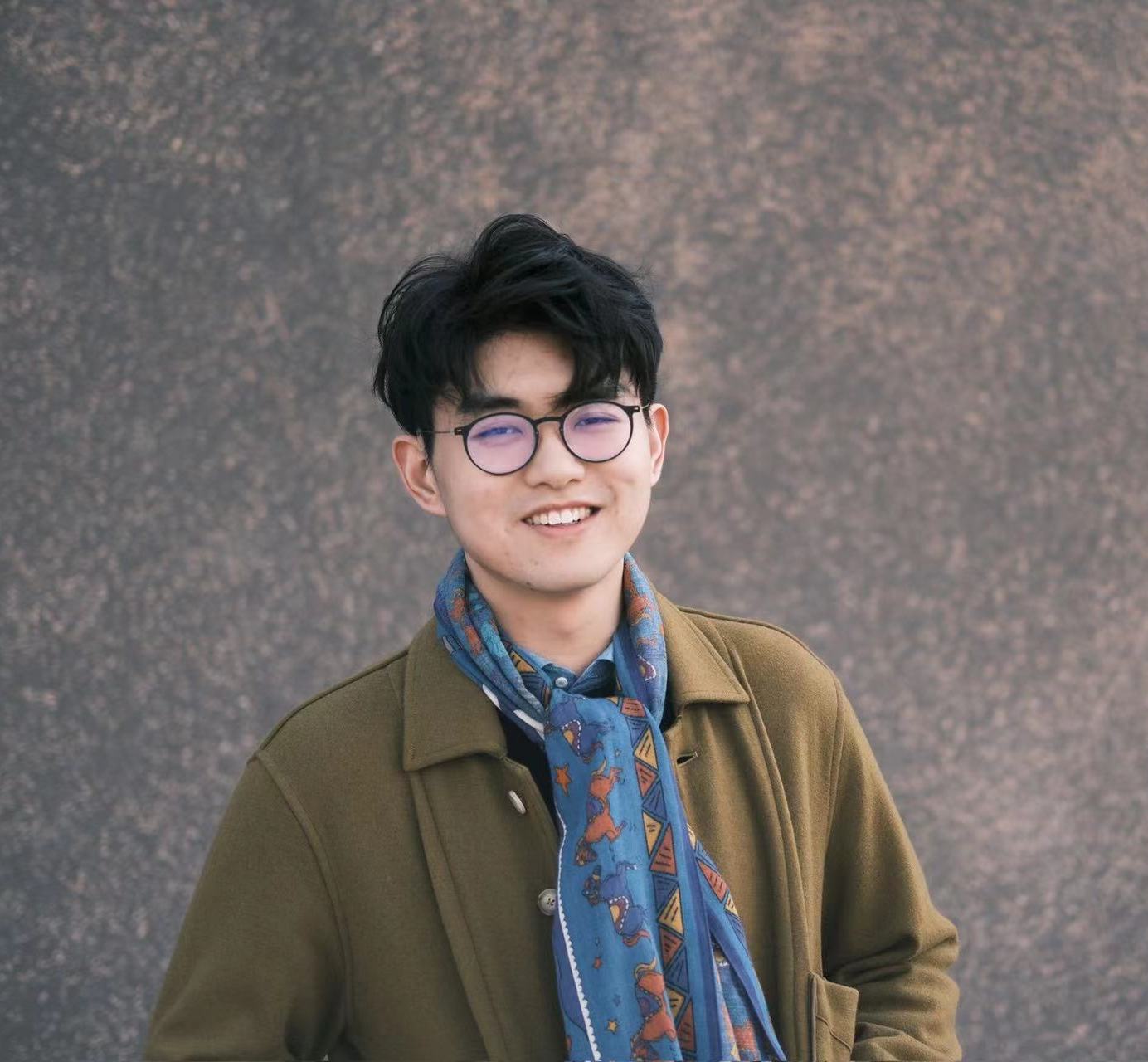}}]
{Yukang Cao} is currently a Research Fellow at MMLab@NTU, Nanyang Technological University, supervised by Prof. Ziwei Liu. 
He received Ph.D degree from the Department of Computer Science, The University of Hong Kong (HKU) advised by Prof. Kwan-Yee K. Wong in 2024.
He was the recipient of HKU-PS scholarship during Ph.D.
He received my B.Eng from Zhejiang University in 2020.
His research interests include computer vision and deep learning.
Particularly, he is interested in 3D representation learning.
\end{IEEEbiography}

\begin{IEEEbiography}
[{\includegraphics[width=1in,height=1.25in,clip,keepaspectratio]{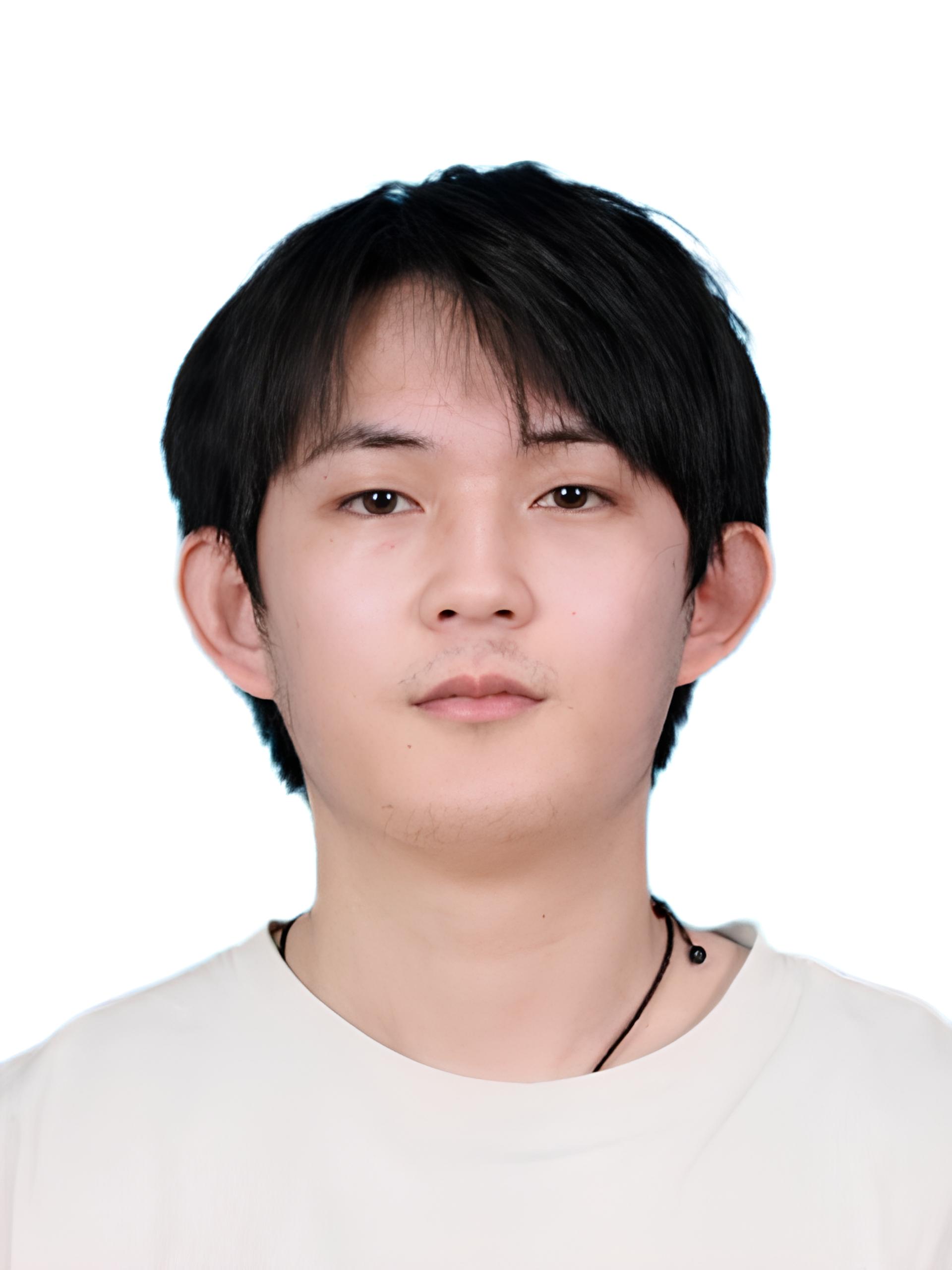}}]
{Jiahao Lu} is currently pursuing a Ph.D. degree at The Hong Kong University of Science and Technology.
He received his bachelor's degree in Artificial Intelligence and Automation from Huazhong University of Science and Technology, Wuhan, Hubei, P. R. China, in 2022. 
His research interests include computer vision and machine learning, with a focus on 3D reconstruction, 3D perception, and 3D generation.
\end{IEEEbiography}

\begin{IEEEbiography}
[{\includegraphics[width=1in,height=1.25in,clip,keepaspectratio]{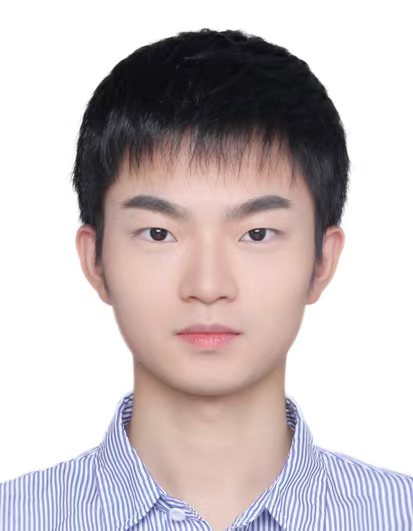}}]
{Zhisheng Huang} is currently a PhD student under the co-supervision of Professor Wenping Wang and Professor Xin Li in the CSE Department at Texas $A\&M$ University (TAMU). 
He completed both his Bachelor's and Master's degrees at Wuhan University. 
His research interests lie in 3D computer vision and graphics.
\end{IEEEbiography}

\begin{IEEEbiography}
[{\includegraphics[width=1in,height=1.25in,clip,keepaspectratio]{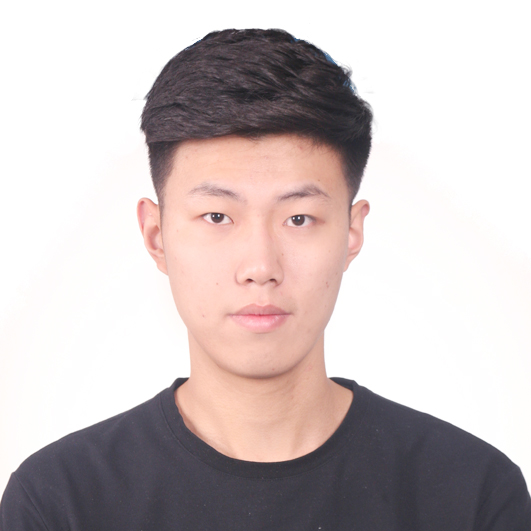}}]
{Zhuowen Shen} is currently pursuing a Ph.D. degree in Computer Science and Engineering at Texas $A\&M$ University, coadvised by Prof. Wenping Wang and Prof. Xin Li. 
He received his Master’s degree in Computer Science and Engineering from the University of Michigan, Ann Arbor, in 2023. 
His research interests lie in computer vision and machine learning, with a focus on 3D reconstruction and 3D representation learning.
\end{IEEEbiography}

\begin{IEEEbiography}
[{\includegraphics[width=1in,height=1.25in,clip,keepaspectratio]{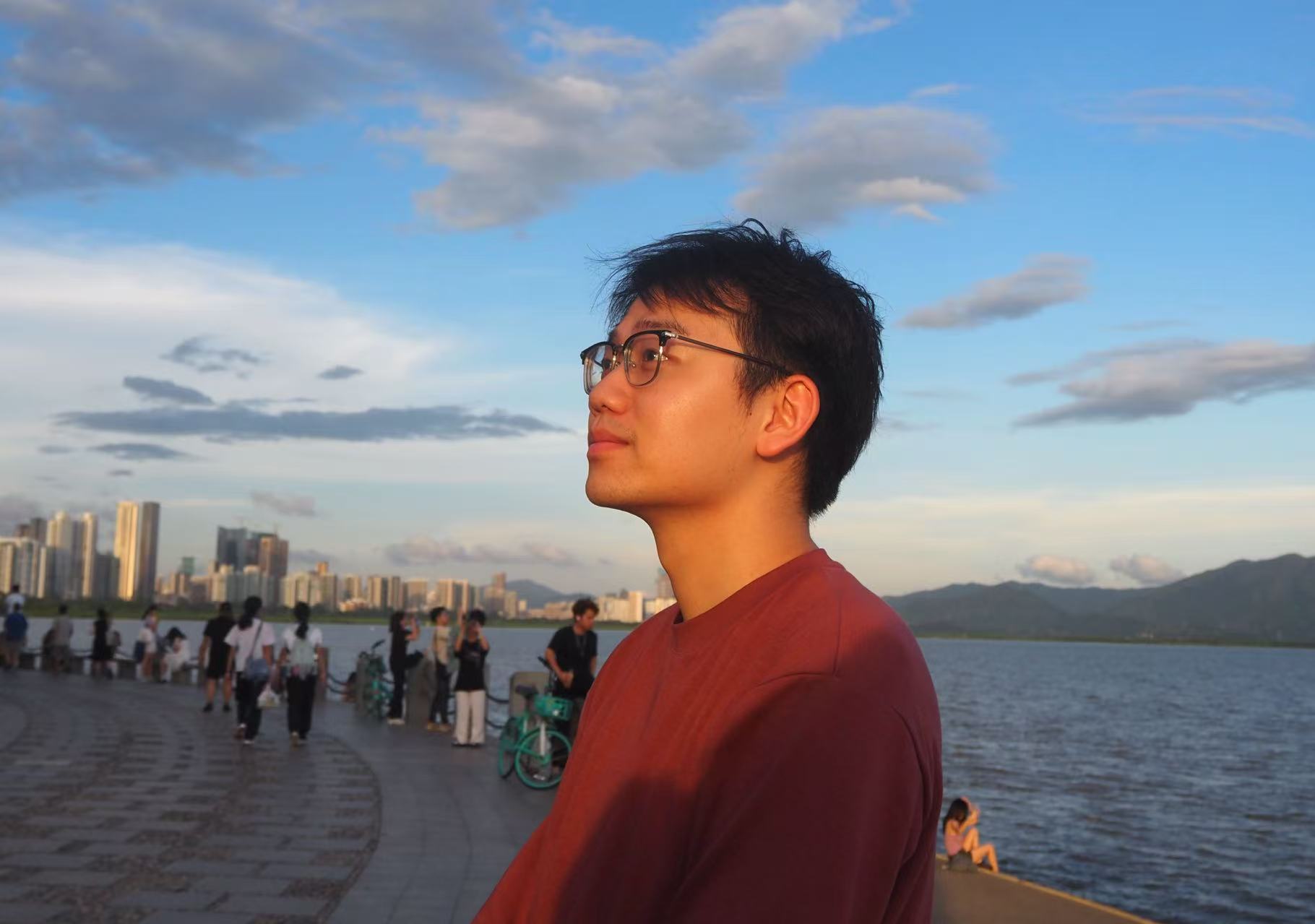}}]
{Chengfeng Zhao} is currently a first-year Ph.D student at Intelligent Graphics Lab in HKUST, supervised by Prof. Yuan Liu. 
Prior to this, He obtained his master and bachelor's degree from ShanghaiTech University, advised by Prof. Lan Xu. 
He was also fortunate to work closely with Prof. Jingyi Yu and Prof. Yuexin Ma.
His research interests are in Computer Graphics and 3D Computer Vision, specifically video generation, human motion synthesis, learning-based garment simulation, large models, etc.
\end{IEEEbiography}

\begin{IEEEbiography}
[{\includegraphics[width=1in,height=1.25in,clip,keepaspectratio]{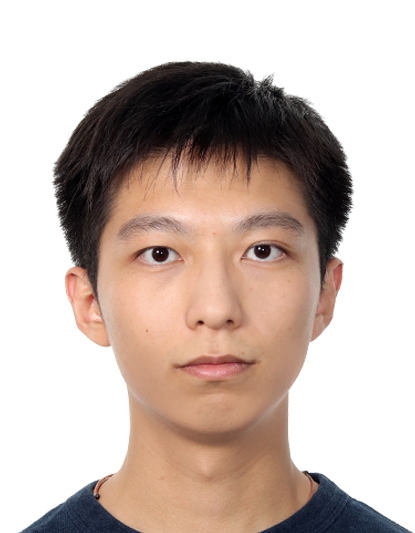}}]
{Fangzhou Hong} is currently a research fellow at MMLab@NTU, Nanyang Technological University, supervised by Prof. Ziwei Liu.
He received Ph.D. degree from MMLab at Nanyang Technological University, supervised by Prof. Ziwei Liu in 2025. He received a B.Eng. degree in software engineering from Tsinghua University, China, in 2020. 
His research interests include computer vision and deep learning. 
Particularly, he is interested in 3D representation learning.
\end{IEEEbiography}

\begin{IEEEbiography}
[{\includegraphics[width=1in,height=1.25in,clip,keepaspectratio]{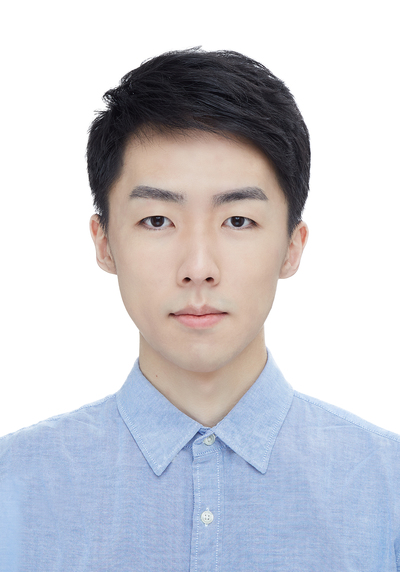}}]
{Zhaoxi Chen} is currently a Ph.D. student at MMLab@NTU, Nanyang Technological University, supervised by Prof. Ziwei Liu.
He received the bachelor’s degree from Tsinghua University, in 2021. 
He received the AISG PhD Fellowship in 2021. 
His research interests include inverse rendering and 3D generative models. 
He has published several papers in CVPR, ICCV, ECCV, ICLR, NeurIPS, TOG, and TPAMI.
He also served as a reviewer for CVPR, ICCV, NeurIPS, TOG, and IJCV.
\end{IEEEbiography}

\begin{IEEEbiography}
[{\includegraphics[width=1in,height=1.25in,clip,keepaspectratio]{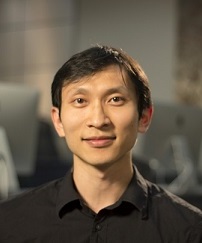}}]
{Xin Li} is currently a Professor and Chair of the Section of Visual Computing and Computational Media, within the College of Performance, Visualization, and Fine Arts.
He is an affiliated faculty member (courtesy appointment) of the Department of Computer Science and Engineering, College of Engineering. 
He is also affiliated with Aggie Computer Graphics Group . 
He got my B.S. degree in Computer Science in 2003 at University of Science and Technology of China (USTC) with a major in Computer Science, and obtained his M.S. and Ph.D. degrees in Computer Science from State University of New York at Stony Brook in 2005 and 2008. 
Before He joined Texas $A\&M$ University, he was a faculty member at School of Electrical Engineering and Computer Science, Louisiana State University (from 2008 to 2022).
\end{IEEEbiography}

\begin{IEEEbiography}
[{\includegraphics[width=1in,height=1.25in,clip,keepaspectratio]{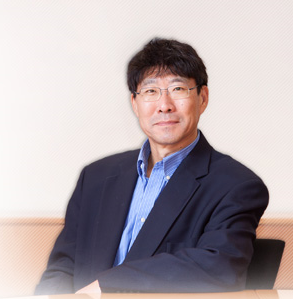}}]
{Wenping Wang} is currently a Professor of Computer Science
$\&$ Engineering at Texas $A\&M$ University. 
His research interests include computer graphics, computer visualization, computer vision, robotics, medical image processing, and geometric computing. 
He has published over 300 technical papers in these fields.
He is journal associate editor of Computer Aided Geometric Design (CAGD) and IEEE Transactions on Visualization and Computer Graphics, and has chaired a number of international conferences, including Pacific Graphics 2012, ACM Symposium on Physical and Solid Modeling (SPM) 2013, SIGGRAPH Asia 2013, and Geometry Summit 2019. 
He received the John Gregory Memorial Award for his contributions in geometric modeling. 
He is an ACM Fellow and IEEE Fellow.
\end{IEEEbiography}

\begin{IEEEbiography}
[{\includegraphics[width=1in,height=1.25in,clip,keepaspectratio]{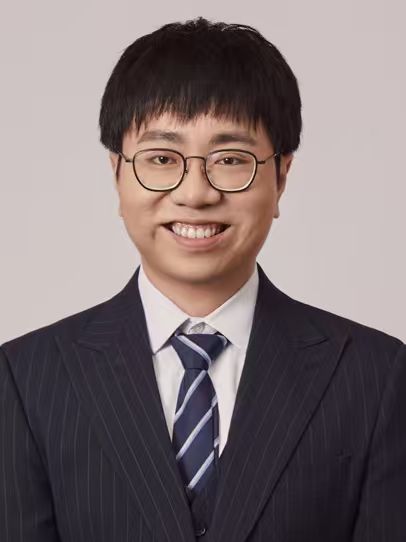}}]
{Yuan Liu} is an assistant professor at the Hong Kong University of Science and Technology (HKUST). 
Prior to that, Yuan worked in Nanyang Technological University (NTU) as a PostDoc researcher and obtained his PhD degree at the University of Hong Kong (HKU). 
His research mainly concentrates on 3D vision and graphics. 
He currently works on topics about 3D AIGC, including 3D neural representations, 3D generative models, and 3D-aware video generation.
\end{IEEEbiography}

\begin{IEEEbiography}
[{\includegraphics[width=1in,height=1.25in,clip,keepaspectratio]{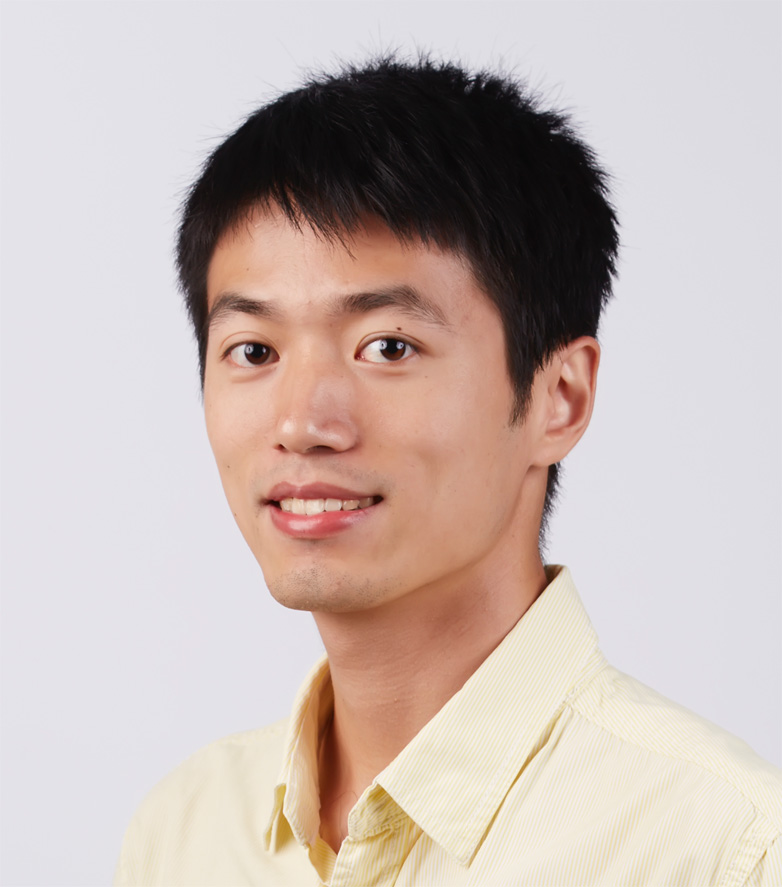}}]
{Ziwei Liu} is currently an associate professor at Nanyang Technological University, Singapore. 
His research revolves around computer vision, machine learning, and computer graphics. 
He has published extensively on top-tier conferences and journals in relevant fields, including CVPR, ICCV, ECCV, NeurlPS, ICLR, ICML, TPAMI, TOG, and Nature Machine Intelligence. 
He is the recipient of the Microsoft Young Fellowship, Hong Kong PhD Fellowship, ICCV Young Researcher Award, HKSTP Best Paper Award and WAIC Yunfan Award. 
He serves as an Area Chair of CVPR, ICCV, NeurlPS, and ICLR, as well as an Associate Editor of IJCV.
\end{IEEEbiography}
\vfill

\end{document}